\documentclass{article} 
\usepackage{arxiv}
\usepackage{url}
\usepackage[utf8]{inputenc}

\usepackage{graphicx}
\usepackage{amsmath}
\usepackage[version=4]{mhchem}
\usepackage{siunitx}
\usepackage{longtable,tabularx}
\usepackage{cleveref}
\usepackage{caption}
\usepackage{subcaption}
\usepackage{algorithm,algpseudocode}
\usepackage[algo2e, linesnumbered]{algorithm2e}
\usepackage{soul}
\usepackage{booktabs}
\SetKwInOut{Input}{Input}
\SetKwInOut{Output}{Output }
\title{Automatic Borescope Damage Assessments for Gas Turbine Blades via Deep Learning}

\author{ Chun Yui Wong \\
	Department of Engineering\\
	University of Cambridge\\
	Cambridge, United Kingdom \\
	\texttt{cyw28@cam.ac.uk} \\
	\And
	Pranay Seshadri \\
	Department of Mathematics (Statistics Section)\\
	Imperial College London\\
	London, United Kingdom \\
	\And
	Geoffrey T. Parks \\
	Department of Engineering\\
	University of Cambridge\\
	Cambridge, United Kingdom \\
}

\begin{document}

\maketitle

\begin{abstract}
To maximise fuel economy, bladed components in aero-engines operate close to material limits. The severe operating environment leads to in-service damage on compressor and turbine blades, having a profound and immediate impact on the performance of the engine. Current methods of blade visual inspection are mainly based on borescope imaging. During these inspections, the sentencing of components under inspection requires significant manual effort, with a lack of systematic approaches to avoid human biases. To perform fast and accurate sentencing, we propose an automatic workflow based on deep learning for detecting damage present on rotor blades using borescope videos. Building upon state-of-the-art methods from computer vision, we show that damage statistics can be presented for each blade in a blade row separately, and demonstrate the workflow on two borescope videos.
\end{abstract}

%\section{Nomenclature}
%
%{\renewcommand\arraystretch{1.0}
%\noindent\begin{longtable*}{@{}l @{\quad=\quad} l@{}}
%$A$  & amplitude of oscillation \\
%$a$ &    cylinder diameter \\
%$C_p$& pressure coefficient \\
%$Cx$ & force coefficient in the \textit{x} direction \\
%$Cy$ & force coefficient in the \textit{y} direction \\
%c   & chord \\
%d$t$ & time step \\
%$Fx$ & $X$ component of the resultant pressure force acting on the vehicle \\
%$Fy$ & $Y$ component of the resultant pressure force acting on the vehicle \\
%$f, g$   & generic functions \\
%$h$  & height \\
%$i$  & time index during navigation \\
%$j$  & waypoint index \\
%$K$  & trailing-edge (TE) nondimensional angular deflection rate
%\end{longtable*}}

\section{Introduction}
Jet engines on aircraft are operated close to material limits to maximise fuel economy. Components of the engine are frequently placed in adverse conditions with strong and fluctuating stresses. Such stresses are exerted by the highly pressurised gases flowing through the engine, irregular impacts from foreign object debris (FOD), and high temperatures at locations such as the output of the combustor. These conditions can cause defects that lead to structural damage of the blades. Due to the adverse operating conditions, even small defects can have a snowball effect and cause catastrophic failure rapidly. In particular, small nicks and bends can upset the balance of rotors, leading to forced vibrations and flutter of blades, drastically reducing the high cycle fatigue life of the entire blade row. In addition, blade surfaces can suffer from corrosion damage. In the case of turbine blades, combustion products contain elements such as sulphur, lead and vanadium which damage the nickel-based alloy that makes up the blade under oxidising conditions; for compressor blades, salt from sea air can form deposits on blades as water evaporates within the compressor, weakening the steel blades via erosion---an effect commonly known as compressor fouling. Works such as \cite{carter2005common} and \cite{meher-homji1998gas} offer detailed qualitative reviews of the effects of commonly found in-service deteriorations in gas turbines.

Owing to the regular occurrence of blade damage, routine inspection of engine components is a crucial part of maintaining adequate performance standards to ensure the airworthiness of an engine. Currently, such inspections are mainly carried out visually via the identification of defects on the components. Borescope imaging is a popular technique that allows an inspector to probe the interior of an engine; the borescope's flexible tube is designed to fit into engine access ports, permitting the inspection to be carried out with minimal disassembly. For example, special access plugs or the hole of a removed igniter can be used to access the hot section of a turbine \cite{federal_aviation_administration2013aviation}. The video feed captured by borescopes is analysed in real time by inspectors to identify potential anomalies. Once a severe flaw is identified, the engine is sent to a maintenance, repair and overhaul (MRO) facility where further inspection is performed by disassembly of the engine. This amounts to an expensive procedure that represents a considerable fraction of the engine list price \cite{aust2019taxonomy}. For example, Boeing reports that the cost of repairing damage caused by FOD can exceed \$1 million for a single overhaul operation per engine, exceeding 20\% of the engine's list price\footnote[2]{https://www.boeing.com/commercial/aeromagazine/aero\_01/textonly/s01txt.html}. Therefore, the sentencing of an engine must be performed prudently but quickly to minimise downtime. 

The process of engine inspection involves a significant investment of manual effort. Inspectors are required to undergo rigorous and costly training. New and inexperienced inspectors may lack the knowledge and skills to reliably identify and flag problematic engine parts \cite{drury2001human}. Human factors present a significant challenge to the maintenance process; the decision of whether an identified defect is deemed deleterious enough to warrant repair operations---or is acceptable and simply carried forward to the next inspection---relies on manual measurements of the location and size of the defect. These measurements are compared against tolerance limits based on previous case studies delineated in relevant engine manuals \cite{tanner2005method}. Tolerance guidelines differ across various components and the type of damage inflicted on the blade---a nick and a score of the same dimension on the same blade can have different impacts on the functionality of the engine, depending on component loads. Various studies have been conducted on quantifying the impact of damaged blades on component \cite{taylor2020predicting,hanschke2017effect} and overall engine performance \cite{khani2012towards,adamczuk2013early}, but have restricted their analyses to specific types of defects, with varying methods of measuring output performance leading to different conclusions. To complicate matters, in \cite{aust2019taxonomy} the authors find that the classification of defects on gas turbine blades presented from various sources is inconsistent and sometimes incomplete, leading to variability in practices among different parties even within a single organisation.

\begin{figure}
\centering
\includegraphics[width=1\linewidth]{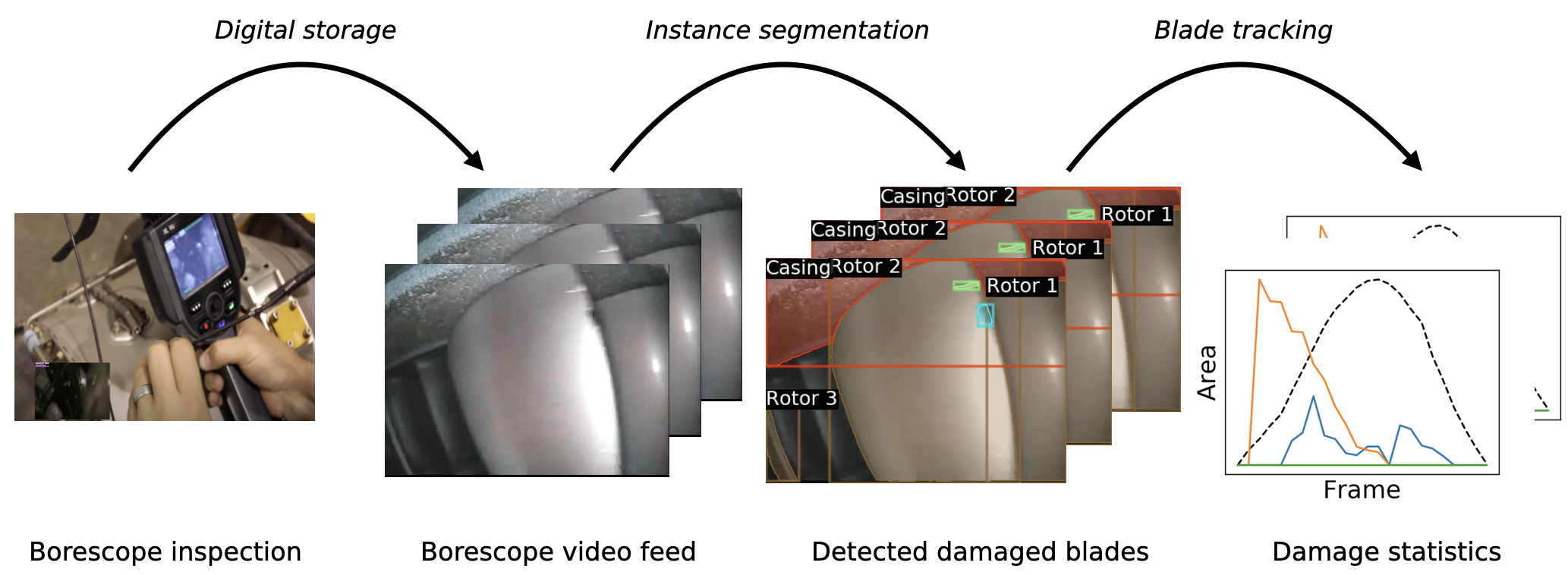}
\caption{Overview of automatic borescope damage assessment workflow. The borescope inspection image on the left is from \protect\url{https://youtu.be/gChfhRZ\_pN0}.}
\label{fig:overview}
\end{figure}

\begin{figure}
     \centering
     \begin{subfigure}[b]{0.45\textwidth}
         \centering
         \includegraphics[width=\textwidth]{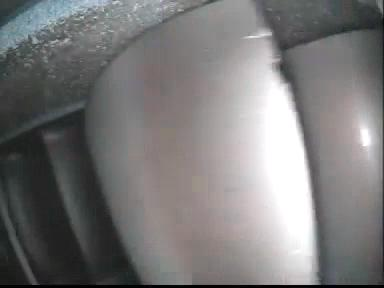}
         \caption{Original image}
     \end{subfigure}
     \hfill
     \begin{subfigure}[b]{0.45\textwidth}
         \centering
         \includegraphics[width=\textwidth]{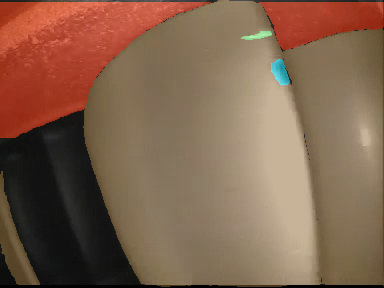}
         \caption{Semantic segmentation}
     \end{subfigure}
     \\
     \vspace{3mm}
     \begin{subfigure}[b]{0.45\textwidth}
         \centering
         \includegraphics[width=\textwidth]{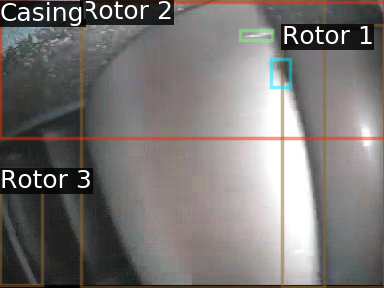}
         \caption{Object detection}
     \end{subfigure}
     \hfill
      \begin{subfigure}[b]{0.45\textwidth}
         \centering
         \includegraphics[width=\textwidth]{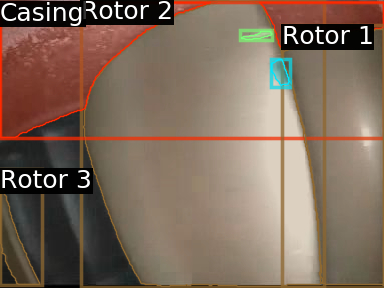}
         \caption{Instance segmentation}
     \end{subfigure}
        \caption{Illustrating different segmentation tasks in computer vision. In the segmentation results, red denotes casing, brown denotes rotor, green denotes surface damage and blue denotes material separation.}
        \label{fig:segm_tasks}
\end{figure}

The aim of the work presented in this paper is to develop an automatic workflow for assessing damage on bladed gas turbine components via analysis of video feeds collected from borescope inspections (see \Cref{fig:overview}). The advantages of an automatic framework include the removal of human biases, a faster decision process, possible gains in accuracy, and---in the face of uncertainty---the ability to quantify this uncertainty via confidence scores. In this workflow, video feeds collected from borescope inspections are stored and input to a bespoke neural network. The structure of the component in the video is deduced via an instance segmentation framework based on deep learning. In instance segmentation, each object within a video frame is detected, classified and localised with a bounding box surrounding the object. Within this bounding box, the object is identified by a mask that highlights relevant pixels. \Cref{fig:segm_tasks} illustrates the differences between three common image segmentation tasks in computer vision, namely semantic segmentation (where each pixel is classified), object detection (where bounding boxes are deduced for each object and classified) and instance segmentation. In literature, neural networks have been deployed to perform semantic segmentation on images collected from borescopes to detect the presence of cracks \cite{shen2019deep} and spallation \cite{bian2016multiscale} on turbine blades. Our work extends and differs from these papers in two main aspects: we aim to (i) classify multiple types of damage, and (ii) deduce the structure of the scene in the borescope video feed by separately identifying multiple blades---noting that, in semantic segmentation, no distinction is made between different instances of the same type of objects. Thus, semantically segmented images are unable to tell us which blade a detected damage region belongs to. By using an instance segmentation framework instead, it is possible to design a procedure to track rotor blades across a sequence of frames as the row is rotated in the captured video during a borescope inspection process. As a result, damage statistics can be extracted for individual blades as they pass through the view of the borescope.

The rest of this paper is structured as follows. In \Cref{sec:method}, we provide an overview of the computational methods used in this work, including a review of methods for instance segmentation and a modular description of the neural network used in our study. In addition, we propose approaches to track detected rotor blades across a sequence of frames, enabling us to create summary statistics of the types and extent of damage found on individual blades. In \Cref{sec:results}, results obtained from the workflow on two example borescope videos are discussed with details of the methods used to assess segmentation quality.

\section{Methodology} \label{sec:method}

\subsection{Instance segmentation with Mask Region-based Convolutional Neural Networks}
Instance segmentation is an active area of research within the computer vision community. Nearly all approaches within this area are based on convolutional neural networks (CNNs) \cite{lecun1998gradient-based,krizhevsky2012imagenet}. These networks are attractive because of their ability to extract features of an image at various scales with translational invariance. Their architecture greatly reduces the number of parameters compared to a fully connected feedforward neural network, enabling greater efficiency in training. Thus, they form the basis of approaches to tackle tasks such as semantic segmentation \cite{long2015fully} and object detection \cite{girshick2014rich,girshick2015fast,ren2017faster,redmon2016you}. Likewise, they are crucial components in state-of-the-art approaches in instance segmentation, including Mask Region-based Convolutional Neural Networks (Mask R-CNN) \cite{he2017mask}, DeepMask \cite{pinheiro2015learning} and Fully Convolutional Instance Segmentation \cite{li2017fully}.

\begin{figure}
\centering
\includegraphics[width=1\linewidth]{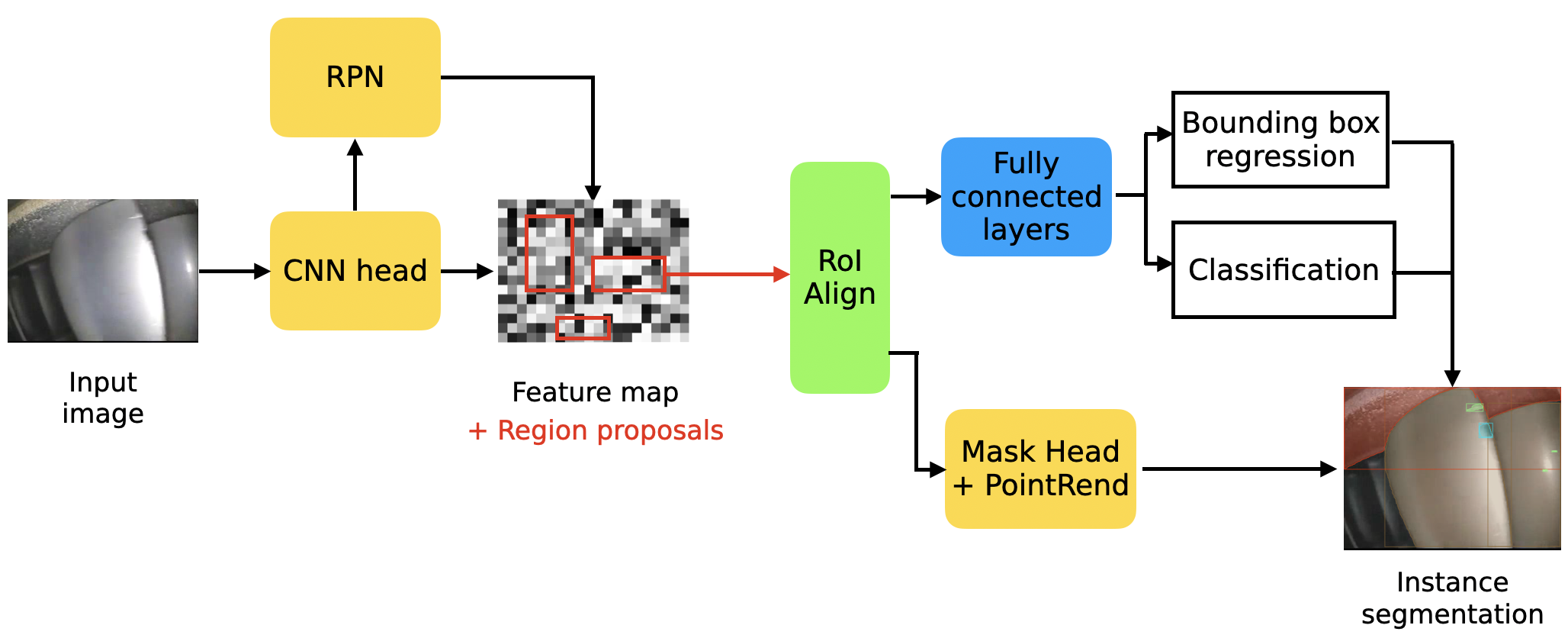}
\caption{Simplified schematic of the Mask R-CNN.}
\label{fig:mask_rcnn}
\end{figure}

In this paper, instance segmentation on contiguous frames of borescope videos is performed by adapting the Mask R-CNN. In the following, we briefly describe the structure of this network, referring readers to the original paper \cite{he2017mask} for further details. The first stage of the Mask R-CNN consists of a CNN head (consisting of convolutional and pooling layers) producing a feature map. This is followed by a Region Proposal Network (RPN), which is another set of convolutional layers that outputs a class prediction and bounding box coordinates for a number of \emph{anchors} centered at each pixel of the feature map. Each anchor is a rectangular box of a certain aspect ratio and scale. The RPN produces a number of region proposals from anchors at different sizes, and assigns them scores indicating the confidence that an object is contained within the proposal bounding boxes. Each proposal, along with the features within the region, is properly aligned into a standard size with a method called RoI Align. Then, it is fed into a network with fully connected layers. This network outputs bounding boxes with refined coordinates that should tightly wrap around objects of interest, along with a classification of the object inside into prescribed classes. Simultaneously, the \emph{mask head} takes in the region proposals and performs semantic segmentation in each region. The architecture of the entire network is shown in \Cref{fig:mask_rcnn} with a simplified representation.

\begin{figure}
\centering
\begin{subfigure}{0.45\textwidth}
\includegraphics[width=1\linewidth]{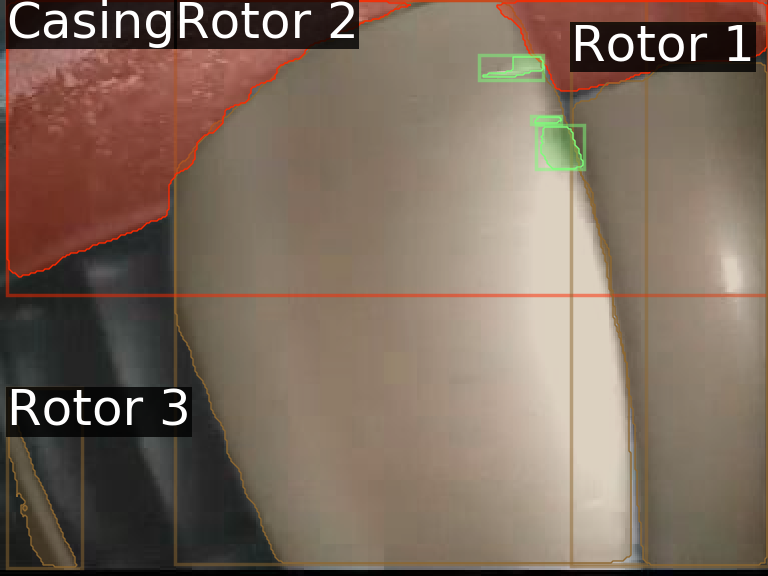}
\caption{Without PointRend}
\label{subfig:no_pr}
\end{subfigure}
\hfill
\begin{subfigure}{0.45\textwidth}
\includegraphics[width=1\linewidth]{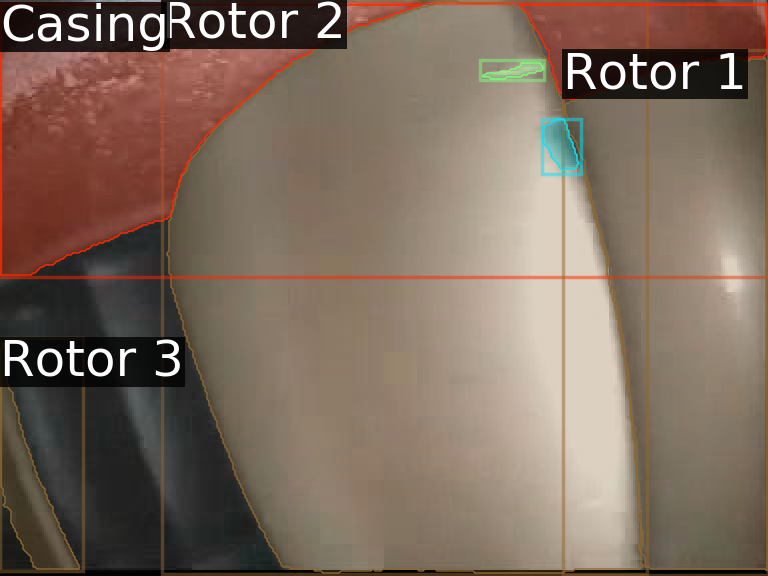}
\caption{With PointRend}
\label{subfig:with_pr}
\end{subfigure}
\caption{Instance segmentation result from Mask R-CNN (a) without and (b) with PointRend. Note the wavy boundary artefacts around the blades and casing in (a).}
\label{fig:old_ins_segm}
\end{figure}

In this project, we find that, on its own, the mask head is unable to produce high quality predictions. As this network is required to operate on region proposals that can differ greatly in scale and aspect ratio within the same image, a network that can effectively segment small objects is unable to produce clean outputs for larger objects, creating undesirable artefacts such as those shown in \Cref{subfig:no_pr}. Therefore, we incorporate the PointRend module \cite{kirillov2020pointrend} on top of Mask R-CNN to improve the segmentation accuracy of the mask head. This module performs point-based segmentation on adaptively re-sampled points on the feature map via an iterative subdivision algorithm. Having more points near the boundary of the object allows finer details of the shape to be captured, thus avoiding the aliasing phenomenon that produces the aforementioned artefacts. As seen in \Cref{subfig:with_pr}, the artefacts are largely removed with the addition of PointRend.

In this work, we adopt the open source software package Detectron2 \cite{wu2019detectron2}, which provides a framework for defining, implementing, training and testing various neural network architectures for computer vision. It is based on the Python deep learning library PyTorch \cite{paszke2019pytorch}, and offers various implementations of Mask R-CNN as well as the PointRend mask head. In this work, the ResNet50-FPN architecture \cite{lin2017feature} is used for the backbone (the part of the network used for feature extraction over the entire image).

\subsection{Blade tracking and calculation of damage statistics}

The Mask R-CNN framework can be deployed to identify and segment objects in individual images. This enables the automatic detection of the presence of different blades in separate frames of the borescope video feed. However, in a contiguous sequence of frames, the same individual blade may be featured in different perspectives, at different positions within the frames. Depending on lighting conditions and the perspective of the borescope camera, various instances of damage may appear in some frames and disappear in others. To provide a holistic assessment of the condition of a certain blade, it is necessary to track blades across the frames they appear in. Since this information is not available from the Mask R-CNN, we design an algorithm to find correlations within a sequence of frames and identify identical blades across adjacent frames.

In \Cref{alg:blade_tracking}, the procedure used to assign unique ID values to blades appearing in a sequence of frames in a video feed is described. Several assumptions are made with this algorithm, which explain the motivation behind its design. First, owing to the contiguous nature of blades moving across frames, we assume that blades do not move significantly across adjacent frames. That is, the displacement of a blade from the previous frame is below a certain threshold. In other words, we assume a high frame rate relative to the rotation speed of the blades. This motivates the approach to match the blade in the current frame to the \emph{nearest} blade in the previous frame in terms of the Euclidean distance between the centres of the respective bounding boxes. For each blade in the current frame, this distance is calculated with all blades in the previous frame. If the minimum distance is below a certain user-set threshold (depending on the video), the blades can be matched and the ID copied onto the current blade. If the minimum distance is larger than the threshold, it is understood that this blade is not seen in the previous frame. Hence, a new ID is set for this blade. In the implementation, we look back to a short sequence of frames (of user-set length $L$) to protect against missing detections in certain frames. Moreover, blade detections with area smaller than a certain user-set threshold are also discarded. In frames with small projected blade areas, it is unlikely that marks of damage on the blade will be clearly visible. We note that the assumption of continuity is broken when the video contains a discontinuity, or the camera is moved too quickly. In practice, some pre-processing is necessary to prevent these possibilities, e.g. by breaking up discontinuous video segments.

{\SetAlgoNoLine%
%\setstretch{1.0}
\begin{algorithm}[H]
\caption{Tracking blades in a video feed.}
\label{alg:blade_tracking}
\begin{algorithm2e}[H]
\Input{Instance segmentation results for each frame, distance, area and detection confidence thresholds, length of lookback $L$. }
\Output {Blade IDs for each blade in each frame.}

\For {each frame} {
Mark all blades with projected area larger than \textit{area threshold} and confidence score above \textit{confidence threshold} as valid.\\
\If {first frame in the video}{
Assign new IDs for each valid blade.\\
\textbf{Continue}\\
}

\For {$k = 1,...,L$} {
\For {each valid unmatched blade in current frame} {
\For {each valid blade $k$ frames ago} {
Calculate the distance between the current blade and previous blade by Euclidean distance between centres of respective bounding boxes.\\
}
Identify blade with minimum distance.\\
\If {minimum distance is smaller than \textit{distance threshold}} {
Set ID of current blade as the ID of the previous blade.\\
}
}
}

\For {each unmatched blade in current frame} {
\uIf {\textit{left-leaving blades} is not empty \textbf{and} current blade's bounding box is on the left side of frame} {
Assign ID as the last left-leaving blade.\\
}
\uElseIf {\textit{right-leaving blades} is not empty \textbf{and} current blade's bounding box is on the right side of frame} {
Assign ID as the last right-leaving blade.\\
}
\Else {
Assign new ID to blade.
}

}

}
\If {blade from previous frame is not found in current frame} {
\eIf {center of bounding box is on the left side of frame} {
Append blade ID to \textit{left-leaving blades}.\\
}
{
Append blade ID to \textit{right-leaving blades}.\\
}
}
\end{algorithm2e}
\end{algorithm}}

Second, we assume that the blades travel (approximately) horizontally, going from left to right or vice-versa. It is possible for the row to travel back and forth, so an unmatched blade may be a blade that has already been seen earlier in the video. To tackle this, lists of blades that have left the view on either side of the travel path are kept, and appropriately assigned to unmatched blades depending on their direction of entry. Clearly, this can be extended to the case of, e.g., vertically travelling blades, but when the borescope is non-stationary during inspection, a more sophisticated method is required.

With a system to track blades across the video, we can collate statistics about the damage on blades as seen from the various frames, identifying which blade the detected damage pixel patches in each frame belongs to. As mentioned earlier, certain marks of damage are more visible in some frames---showing a blade from certain perspectives---than others. Thus, instead of focusing on a single view, damage statistics across the whole trajectory of the blade should be recorded. In \Cref{sec:results}, damage is recorded for each blade as a fraction of the projected area of the blade visible in that frame. We note that this is not a unique method of quantifying and assessing damage; factors other than the area/size of the damage---such as the location of the damage---can be taken into account.

\section{Results} \label{sec:results}

\begin{figure}
\centering
\begin{subfigure}[b]{0.45\textwidth}
\includegraphics[width=1\linewidth]{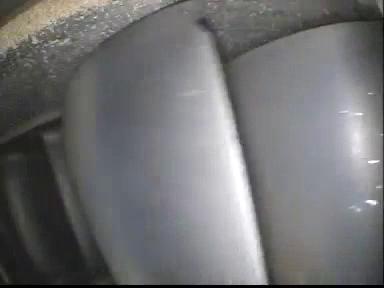}
\caption{Example frame from video 1.}
\label{subfig:ex0}
\end{subfigure}
\hfill
\begin{subfigure}[b]{0.45\textwidth}
\includegraphics[width=1\linewidth]{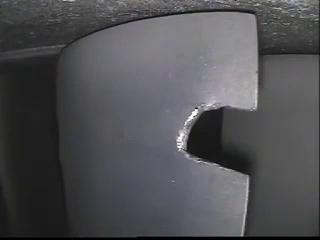}
\caption{Example frame from video 2.}
\label{subfig:ex1}
\end{subfigure}
\caption{Example images from both videos used in this work.}
\label{fig:ex_frames}
\end{figure}

In this section, the workflow described in this paper is demonstrated on borescope images of a high pressure compressor (HPC) collected from two videos available from YouTube\footnote{Video 1: \url{https://youtu.be/eO6GRU4RfC4} and video 2: \url{https://youtu.be/nWk6IeL1IiM}}
 (See \Cref{fig:ex_frames} and \Cref{tab:vids} for example images from and details of both videos.) First, further details on the implementation and deployment of the workflow are provided. Second, performance metrics for the instance segmentation are provided with discussion. Third, we describe a post-processing step to improve detection accuracy. Finally, examples of assessing damage on specific rotor blades are shown including a video visualisation.

\begin{table}
\begin{center}
 \caption{Details of each borescope video used in this work.}
\label{tab:vids}
\begin{tabular}{ lcc } 
\hline \hline
                         & Video 1             & Video 2             \\
 \hline 
Engine                   & \multicolumn{2}{c}{Rolls-Royce RB211-535} \\
Component                & HPC fourth stage    & HPC second stage    \\
Frames per second        & 25                  & 25                  \\
Number of blades visible & 97 (entire row)     & 38 \\
 \hline \hline
 \end{tabular}
 \vspace{3mm}
\end{center}
\end{table}

\subsection{Gathering of training and testing data}
\begin{figure}
\centering
\includegraphics[width=0.5\linewidth]{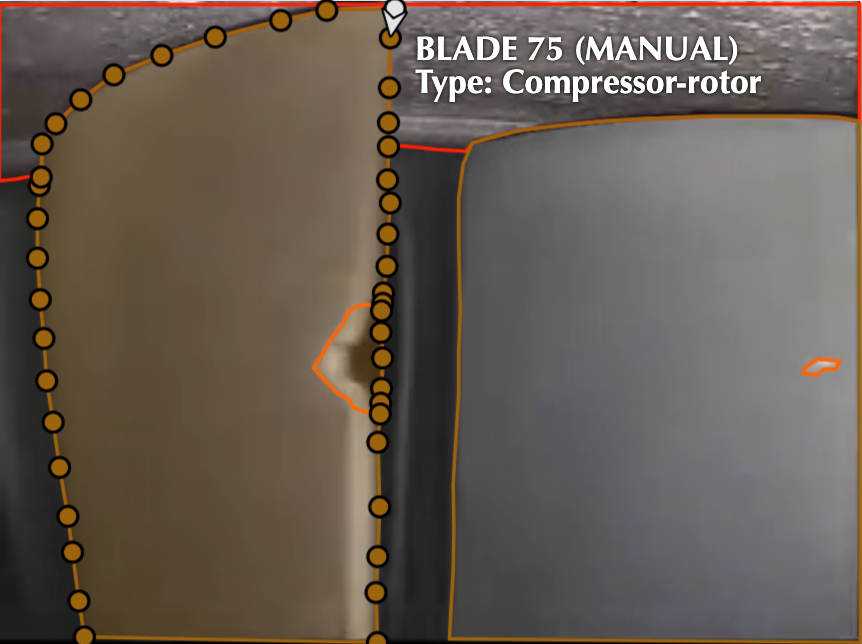}
\caption{Example of labelled image from CVAT.}
\label{fig:example_cvat}
\end{figure}

Out of the 4881 video frames contained in both videos combined, 104 images are provided with manual labels via the open source Computer Vision Annotation Tool (CVAT) software\footnote{\url{https://zenodo.org/record/4009388\#.YDPEKC2l3jA}}. Labels are provided in the form of polygons that approximate the outline of an object (see \Cref{fig:example_cvat}), and are associated with five possible classes, namely \texttt{Casing}, \texttt{Compressor-rotor}, \texttt{Surface-damage}, \texttt{Material-separation} and \texttt{Material-deformation}. In this dataset, \texttt{Surface-damage} mainly refers to scratches, but can also refer to compressor fouling, turbine sulfidation etc. \texttt{Material-separation} includes nicks, tears and cracks; and \texttt{Material-deformation} commonly refers to bends. With reference to \cite{aust2019taxonomy}, these categories are devised to broadly separate forms of damage of different nature (hence different potential impact on performance), while keeping the number of classes small.

It is important to note the introduction of possible human biases in this process. First, labels may not completely tightly wrap around the object of interest. Due to the limited resolution of video frames and the presence of imaging artefacts, this is difficult to avoid, especially when concerned with small damage marks. Second, material separation and deformation damage cause the blade shape to warp. Where they occur, we interpret the blade shape as that corresponding to the original undamaged geometry. Thus, labelling blades with these types of damage involves interpolation/extrapolation of certain points, which can suffer from minor variations. Despite the above, we argue that overall, the automatic framework should mitigate the impact of possible biases compared to a fully manual procedure. For example, the system can highlight damage that may be missed in manual inspections.

\subsection{Training of the neural network}
From the 104 labelled images, 89 images are used for training, and 15 for testing. Network weights are initialised from the pre-trained model for the COCO dataset \cite{lin2014microsoft} \footnote{Weights are available at \url{https://dl.fbaipublicfiles.com/detectron2/PointRend/InstanceSegmentation/pointrend_rcnn_R_50_FPN_3x_coco/164955410/model_final_edd263.pkl}}. The network is trained with a learning rate of 0.02, a weight decay parameter of 0.0001 and momentum of 0.9 using stochastic gradient descent (SGD). The mini-batch size is 16, distributed across 4 GPUs. The network is trained for 2000 iterations (see the next subsection for discussion of this choice). Most of the network configurations follow the defaults set in \cite{wu2019detectron2} according to \cite{he2017mask} and \cite{kirillov2020pointrend}, except for the anchor sizes. Due to the lower resolution of our images ($384 \times 288$), and the presence of small surface damage marks, the proposal anchor sizes for the RPN are set as 16, 32, 64, 128 and 256, smaller than the default.

The multi-task loss function of the Mask R-CNN is composed of five terms,
\begin{equation} \label{eqn:losses}
L_{tot} = L_{rpn, cls} + L_{rpn, reg} + L_{cls} + L_{reg} + L_{mask}.
\end{equation}
The first two terms refer to losses of the region proposals generated by the RPN. The first term $L_{rpn, cls}$ is a binary cross entropy loss over whether a region proposal contains or does not contain an object. The term $L_{rpn, reg}$ penalises discrepancies of the bounding box with ground truth bounding boxes automatically generated from the polygon labels, and is only calculated for \emph{positive} region proposals, i.e. those with sufficient overlap with a ground truth box. The latter three terms refer to the mask head. The term $L_{cls}$ is a \emph{multinomial} cross entropy loss, now penalising wrong classifications of the object within a bounding box. $L_{reg}$ works similarly to $L_{rpn,reg}$ but for final output bounding boxes instead of region proposals. Finally, $L_{mask}$ is the semantic segmentation loss within each positive bounding box.
%,
%\begin{equation}
%L_{cls} = -\sum_{n=1}^N \sum_{c=1}^C t_c^{(n)} \log \left(p^{(n)}_c\right) w_c,
%\end{equation}
%where $N$ is the number of positive region proposals from the RPN, $C$ is the number of classes, $t_c$ is the ground truth class label, and $p_c$ is the predicted confidence value. In this work, we have added class weights $w_c$ to weigh the loss in favour of damage classes, to improve the detection of damage regions. This is because of the small size of damage pixel patches, and the difficulty of distinguishing between different types of damage compared to, say, distinguishing between a blade and the casing. The term $L_{reg}$ is similar, but is now class-aware and only relevant to positive boxes that correctly predict the class labels.

We note that the number of training iterations and number of training images is small compared with common benchmark datasets for instance segmentation (e.g. COCO \cite{lin2014microsoft}, Cityscapes \cite{cordts2016cityscapes}). However, it is shown in the following that this nevertheless results in satisfactory quality in the predictions. This is mainly because of the similarity of the frames in a single video feed. In practice, if the scope of the workflow were to be increased by sourcing training data from borescope inspections of more distinct components, the size of the training set and duration of training would need to be increased appropriately.

\subsection{Results on testing data}

\begin{figure}
\centering
\begin{subfigure}[b]{0.49\textwidth}
\includegraphics[width=1\linewidth]{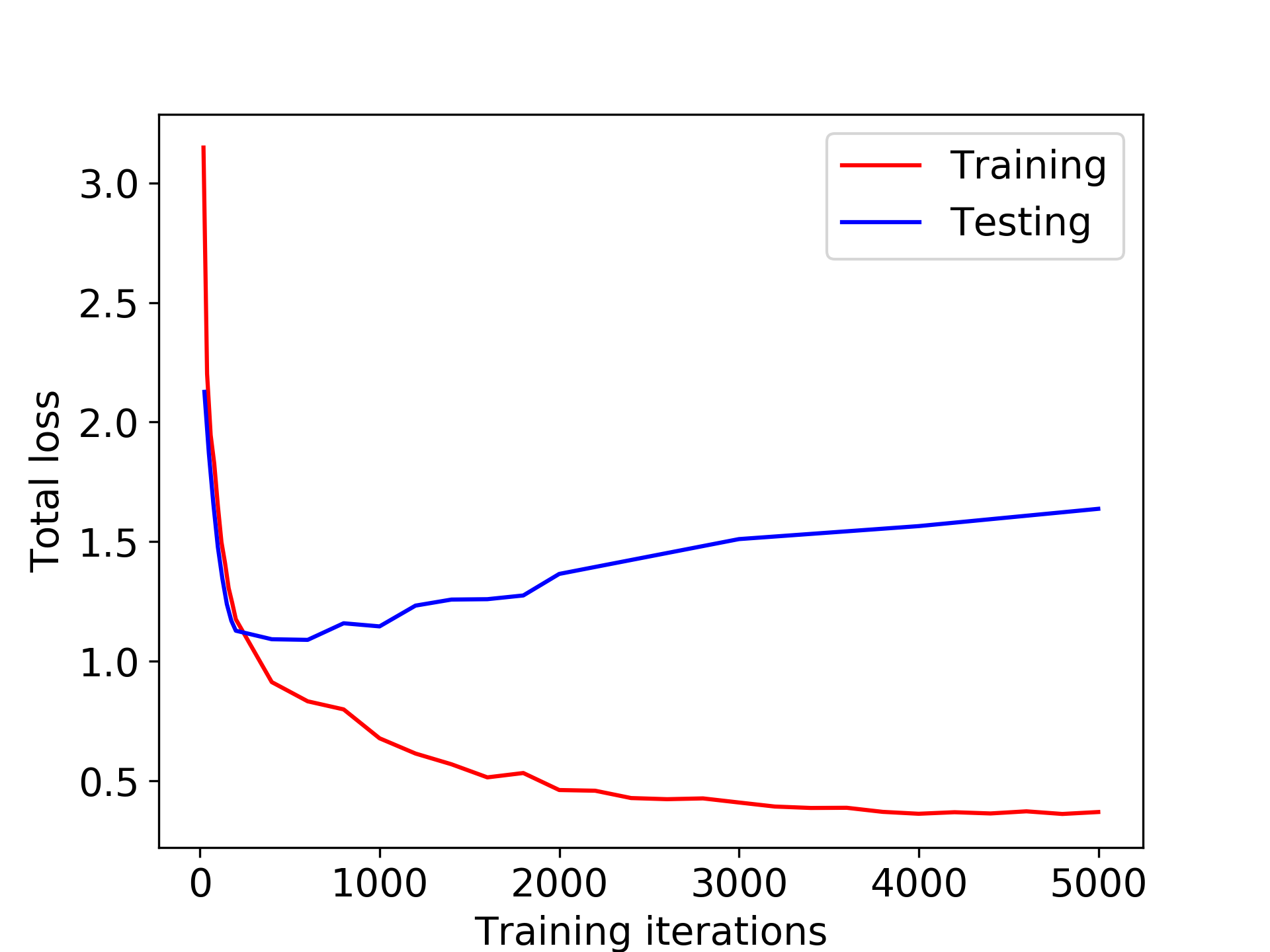}
\caption{}
\label{subfig:losses}
\end{subfigure}
\begin{subfigure}[b]{0.49\textwidth}
\includegraphics[width=1\linewidth]{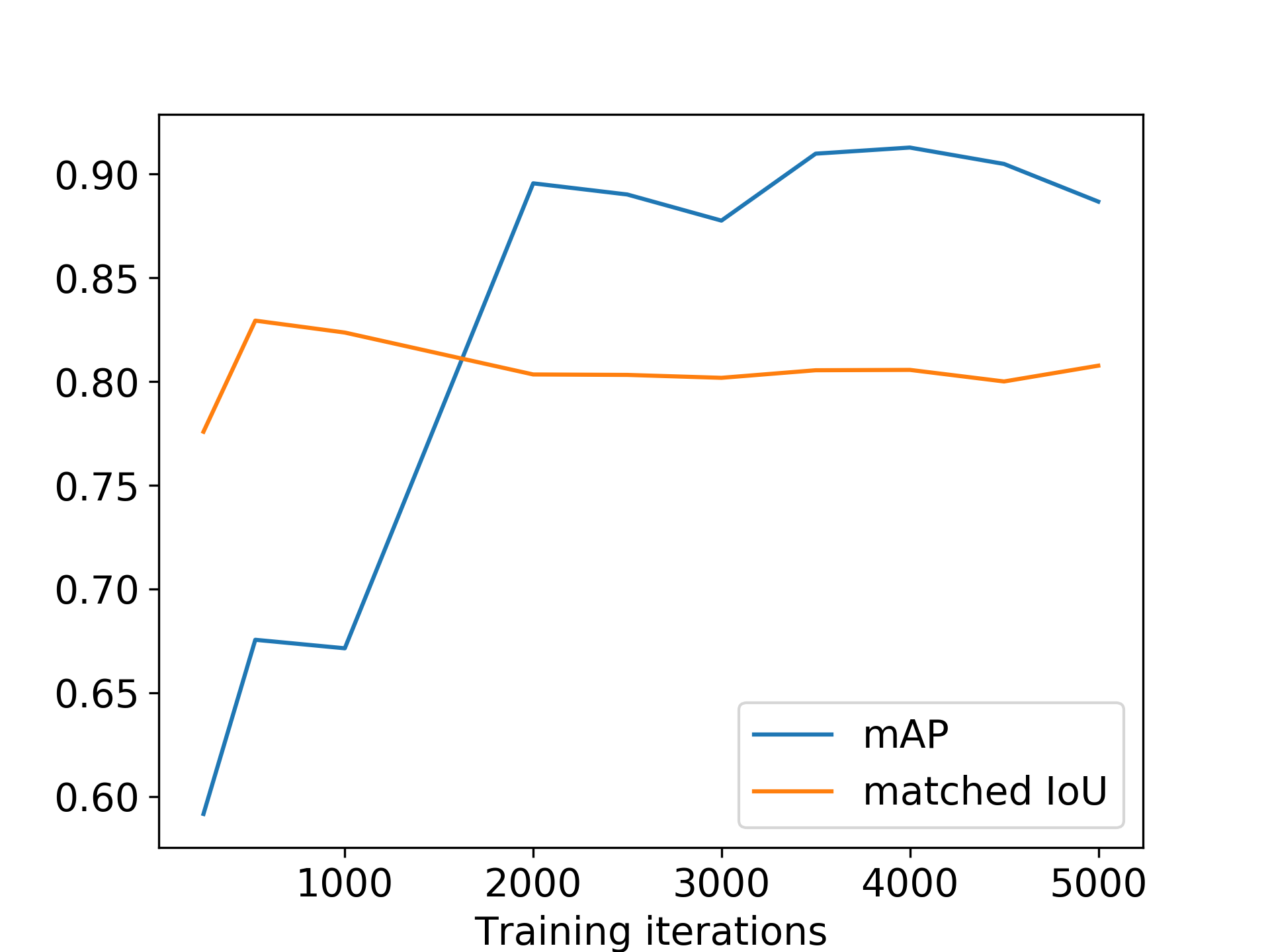}
\caption{}
\label{subfig:mAP}
\end{subfigure}
\caption{(a) Training and testing losses, and (b) mean average precision (mAP) and mean matched IoU on testing data with respect to training iterations.}
\end{figure}

\begin{figure}
\centering
\begin{subfigure}[b]{0.33\textwidth}
\includegraphics[width=1\linewidth]{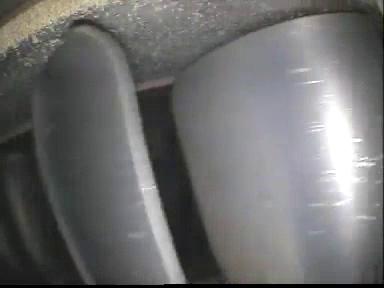}
\caption{Original image}
\label{subfig:frame_1414}
\end{subfigure}
\begin{subfigure}[b]{0.33\textwidth}
\includegraphics[width=1\linewidth]{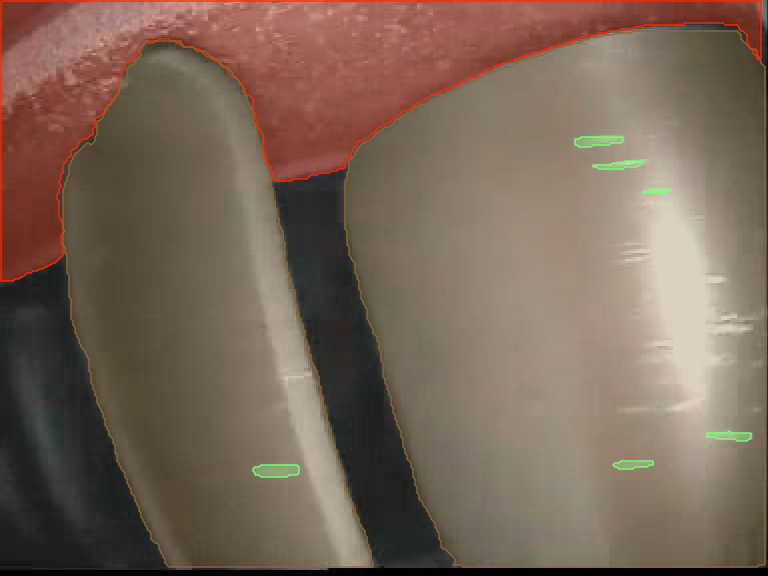}
\caption{Segmentation after 500 iterations}
\label{subfig:frame_1414_525}
\end{subfigure}
\begin{subfigure}[b]{0.33\textwidth}
\includegraphics[width=1\linewidth]{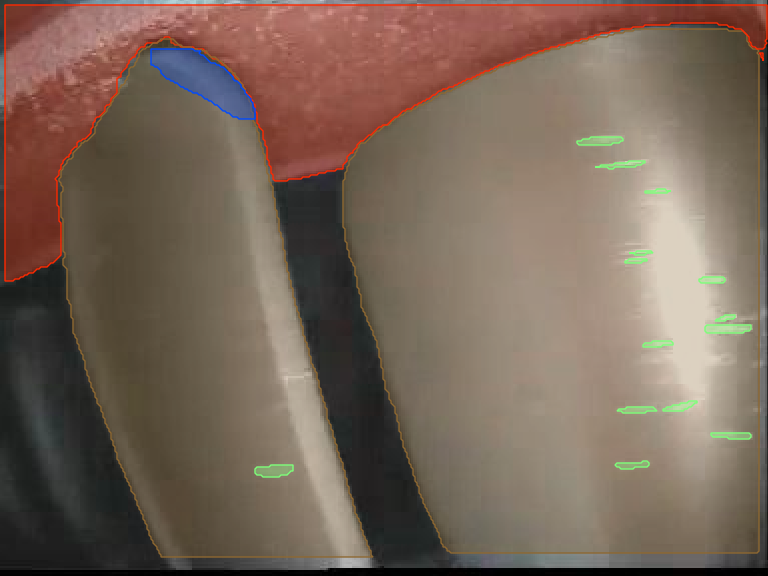}
\caption{Segmentation after 2000 iterations}
\label{subfig:frame_1414_2000}
\end{subfigure}
\caption{Segmentation results for a frame showing rotors 46 and 47 after 500 and 2000 training iterations. Note that this frame is not in the training data.}
\label{fig:frame_1414}
\end{figure}

In \Cref{subfig:losses}, the total loss \eqref{eqn:losses} is plotted against the number of training iterations. Although the loss on the training data continues decreasing as the number of iterations increase, the loss on the testing data increases after about 200 iterations. This suggests that the model is overfitting past 200 iterations. However, it is observed that detection quality is improved when comparing the network at a later stage to that at an earlier stage of training, even on unseen testing images. For example, \Cref{fig:frame_1414} shows a testing image and the corresponding network predictions after 500 and 2000 training iterations respectively. This figure shows that the network after 500 iterations tends to skip the identification of certain damage marks---e.g. some surface damage on rotor 46 and the tip bend near the leading edge of rotor 47. Therefore, the testing loss is not an adequate metric for judging the quality of the network in this case. A possible cause of this is that the loss does not penalise missing detections severely enough---e.g. the mask segmentation loss $L_{mask}$ is only defined on valid region proposals.

Thus, we propose to evaluate the network using two different metrics. They are based on comparing predicted masks with masks from the ground truth (manually labelled images) using the intersection over union (IoU) metric. For two masks $A$ and $B$, their IoU is
\begin{equation} \label{eqn:IoU}
IoU = \frac{\text{Area}(A \cap B)}{\text{Area}(A\cup B)},
\end{equation}
which is a measure of how much overlap the masks have. The mean average precision (mAP) is a common metric used in computer vision literature to evaluate instance segmentation networks. It is evaluated using the steps detailed in \Cref{alg:mAP} (also see \cite{padilla2020survey}) at a certain IoU threshold that we set at 0.5. This is also known as AP$_{50}$ in literature \cite{he2017mask}. It is a measure of the proportion of masks that correctly predict objects in the ground truth of an image, in terms of both the class and a minimum IoU. In addition to this metric, we devise a similar metric---the matched IoU--- that measures the average IoU of True mask predictions with their assigned ground truth masks obtained from \Cref{alg:mAP} for each class. This is subsequently averaged across all classes. This metric measures the \emph{quality} of matched masks in addition to their well-matching with ground truth masks, since the calculation of areas is a crucial part of our results. In \Cref{subfig:mAP}, we have plotted these two metrics averaged across the images in the testing data, evaluated for the network trained at different numbers of training iterations. From this figure, it is clear that the network performs better from  2000 iterations onwards. While the matched IoU stays relatively constant, the number of missing detections is drastically reduced as we train the network, raising the mAP score.

{\SetAlgoNoLine%
%\setstretch{1.0}
\begin{algorithm}[H]
\caption{Calculating mAP for an image.}
\label{alg:mAP}
\begin{algorithm2e}[H]
\Input{Masks and associated class predictions and ground truth, IoU threshold value between 0 and 1.}
\Output {mAP value.}
\For {each class} {
List all predicted masks of a certain class, ordered in descending order of prediction confidence. \\
\For {each predicted mask in descending confidence} {
\If {no unassigned ground truth mask of the same class exists} {
Mark this prediction as False.\\
}
Find the unassigned ground truth mask of the same class that has the highest IoU with the prediction. \\
\If {this IoU $>$ IoU threshold} {
Mark this prediction as True.\\
Mark this ground truth mask as assigned.\\
}
\Else {
Mark this prediction as False.\\
}
}
Set $FN = $ number of unassigned ground truth masks.\\

\For {each predicted mask in descending confidence} {
Calculate $\tilde{P} = $ the number of True predictions so far divided by the number of predictions so far.\\
Calculate $R = $ the number of True predictions so far divided by (the total number of True predictions + $FN$).\\
}
Associate each $\tilde{P}$ with the corresponding $R$ to form $\tilde{P}(R)$, the precision-recall curve.\\
Calculate $P(R) = \text{max}_{R' \geq R} \tilde{P}(R')$, the interpolated precision-recall curve.\\
Calculate $AP = \int_0^1 P(R) dR$.
}
Calculate $mAP = $ the mean of $AP$ across all classes.
\end{algorithm2e}
\end{algorithm}}

\begin{figure}
\centering
\includegraphics[width=1\linewidth]{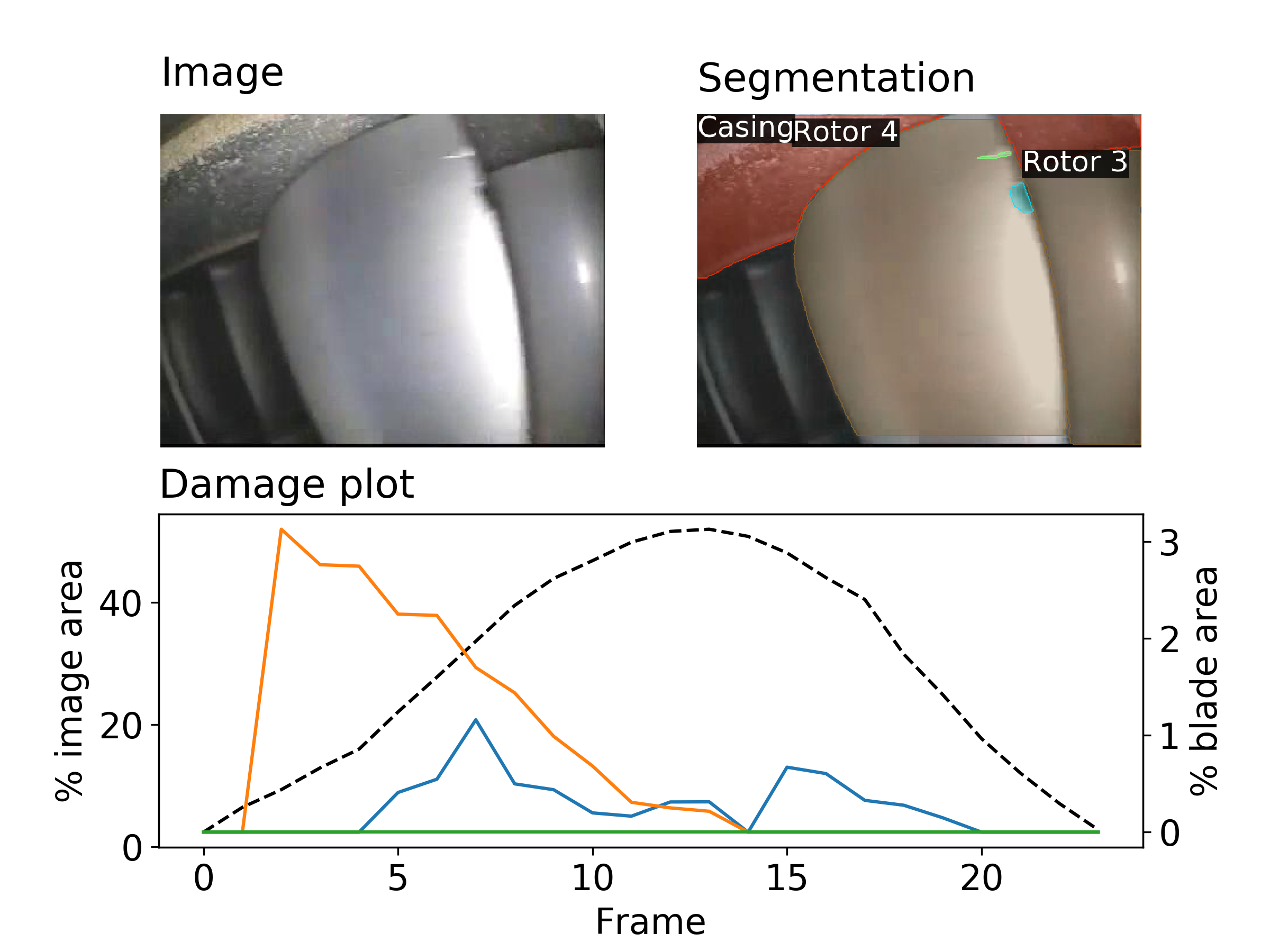}
\caption{Example image, corresponding segmentation result and damage time series plot for Rotor 4, video 1.}
\label{fig:bigplot0}
\end{figure}

\begin{figure}
\centering
\includegraphics[width=1\linewidth]{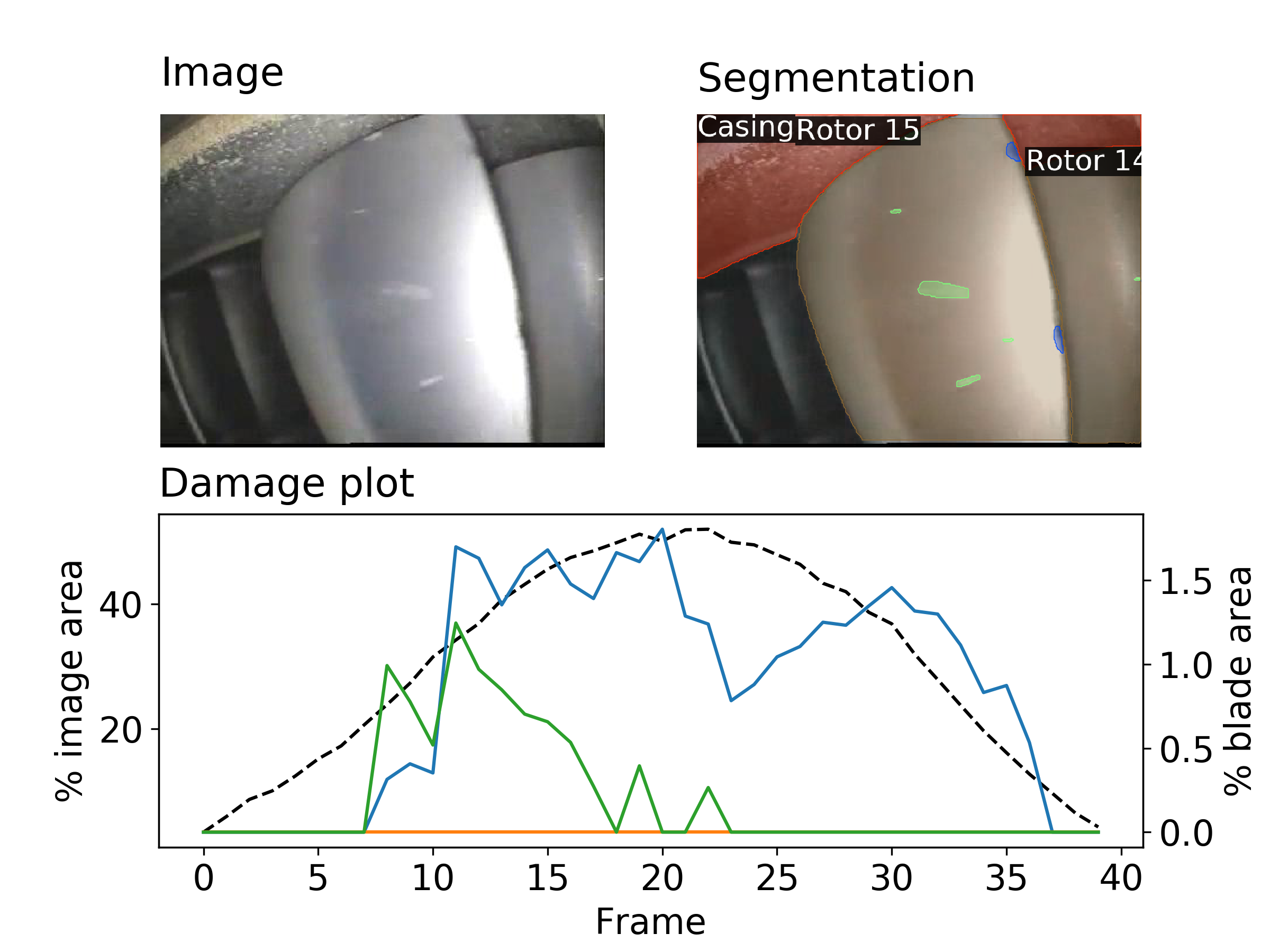}
\caption{Example image, corresponding segmentation result and damage time series plot for Rotor 15, video 1.}
\label{fig:bigplot1}
\end{figure}

\begin{figure}
\centering
\includegraphics[width=1\linewidth]{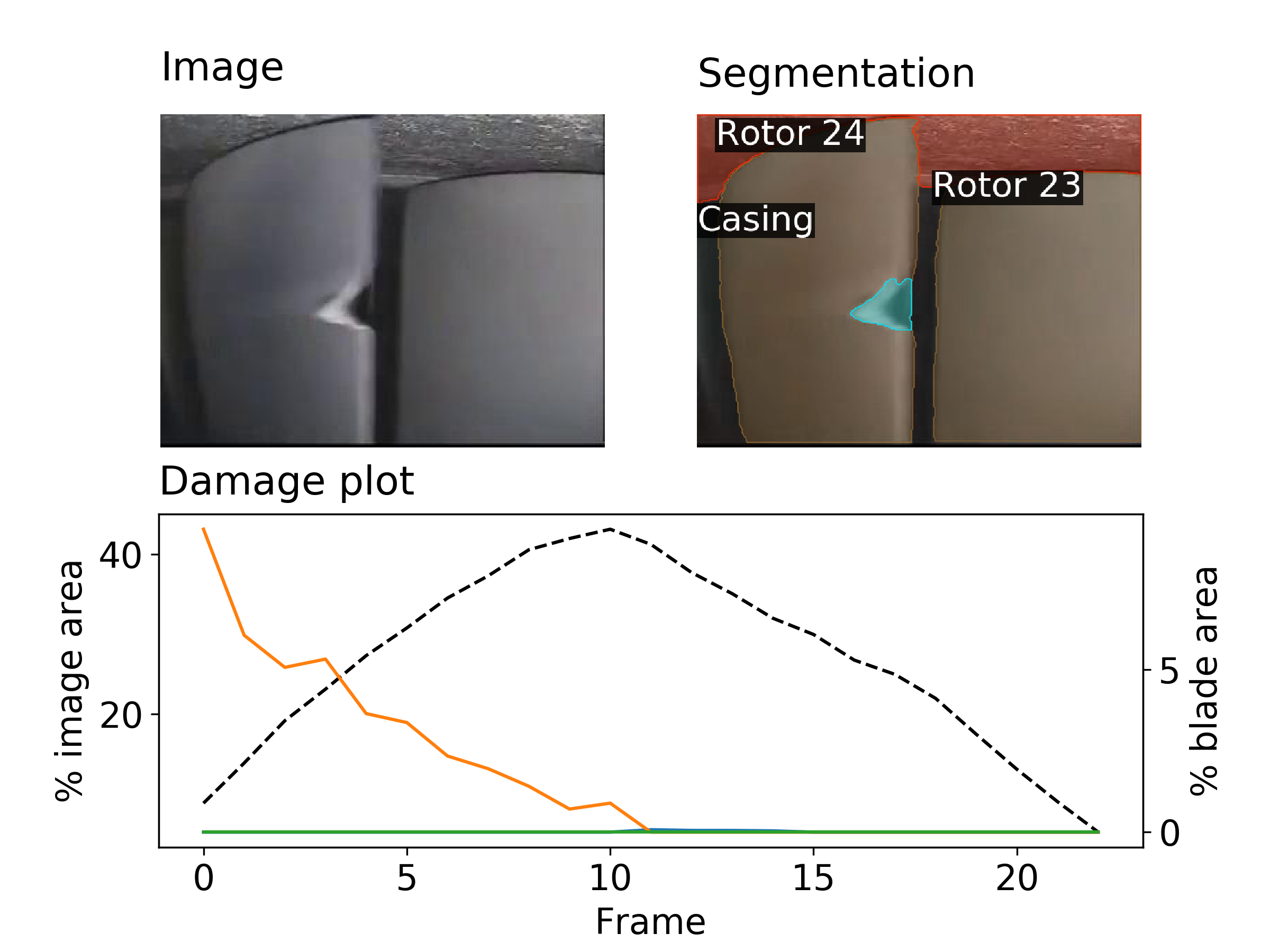}
\caption{Example image, corresponding segmentation result and damage time series plot for Rotor 24, video 2.}
\label{fig:bigplot2}
\end{figure}

Having obtained the segmentation of each frame in the video, \Cref{alg:blade_tracking} is applied to correlate the masks across adjacent frames and assign unique IDs to each rotor found within the video sequence. Then, the projected areas of the rotors shown in the images, along with the associated area of damage present on the blades, are calculated by counting the pixels in their masks. For each rotor, it is possible to plot a time series graph, showing the amount of damage present on each blade relative to the blade projected area as the blade emerges into and then leaves the camera view. In \Cref{fig:bigplot0,fig:bigplot1,fig:bigplot2}, we show three example time series plots of the detected damage on rotors. In all three plots, the black dashed line shows the detected rotor area as a fraction of image area, and the coloured solid lines show different types of damage as a fraction of blade area; the blue line shows surface damage, the orange line material separation and green line material deformation. As the rotors travel across the camera view, the extent of detected damage varies. Therefore, a thorough assessment of rotor damage should take into account the full time series.

\subsection{Post-processing to identify surface damage}
\begin{figure}
\centering
\begin{subfigure}[b]{0.49\textwidth}
\centering
\includegraphics[width=0.8\linewidth]{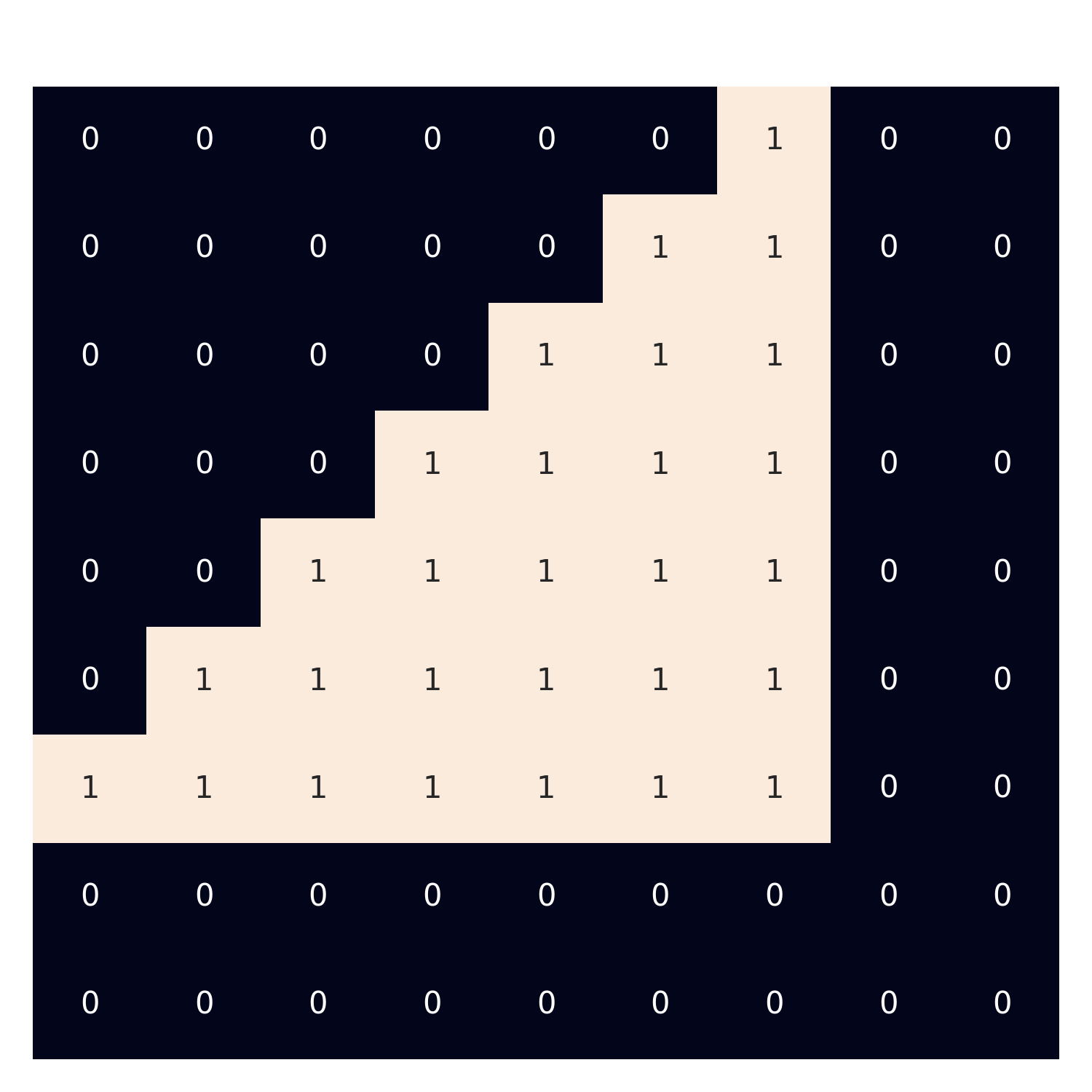}
\caption{Before erosion}
\end{subfigure}
\begin{subfigure}[b]{0.49\textwidth}
\centering
\includegraphics[width=0.8\linewidth]{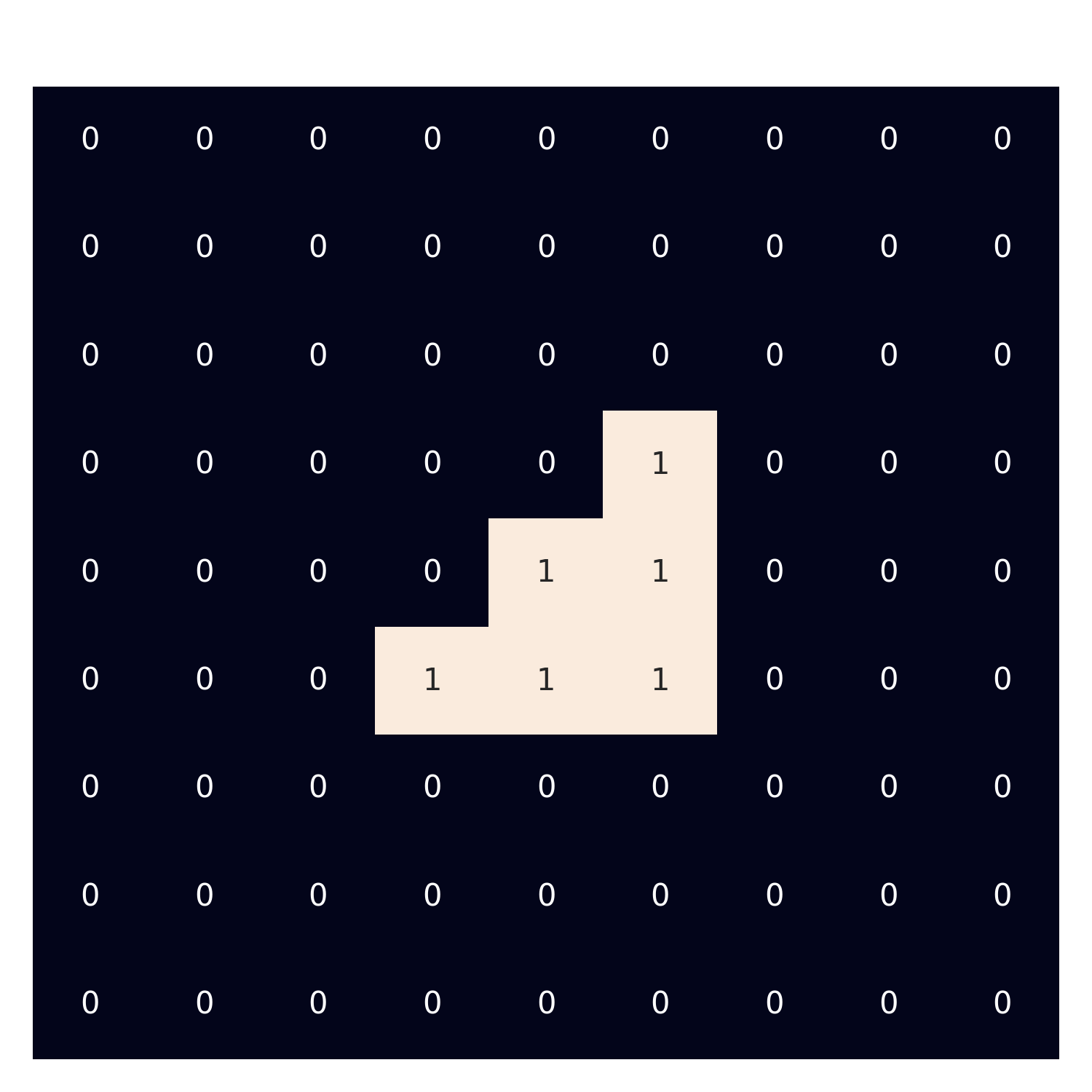}
\caption{After erosion}
\end{subfigure}
\caption{Demonstrating the morphological erosion operator on an example binary array.}
\label{fig:erosion}
\end{figure}

\begin{figure}
\centering
\includegraphics[width=1\linewidth]{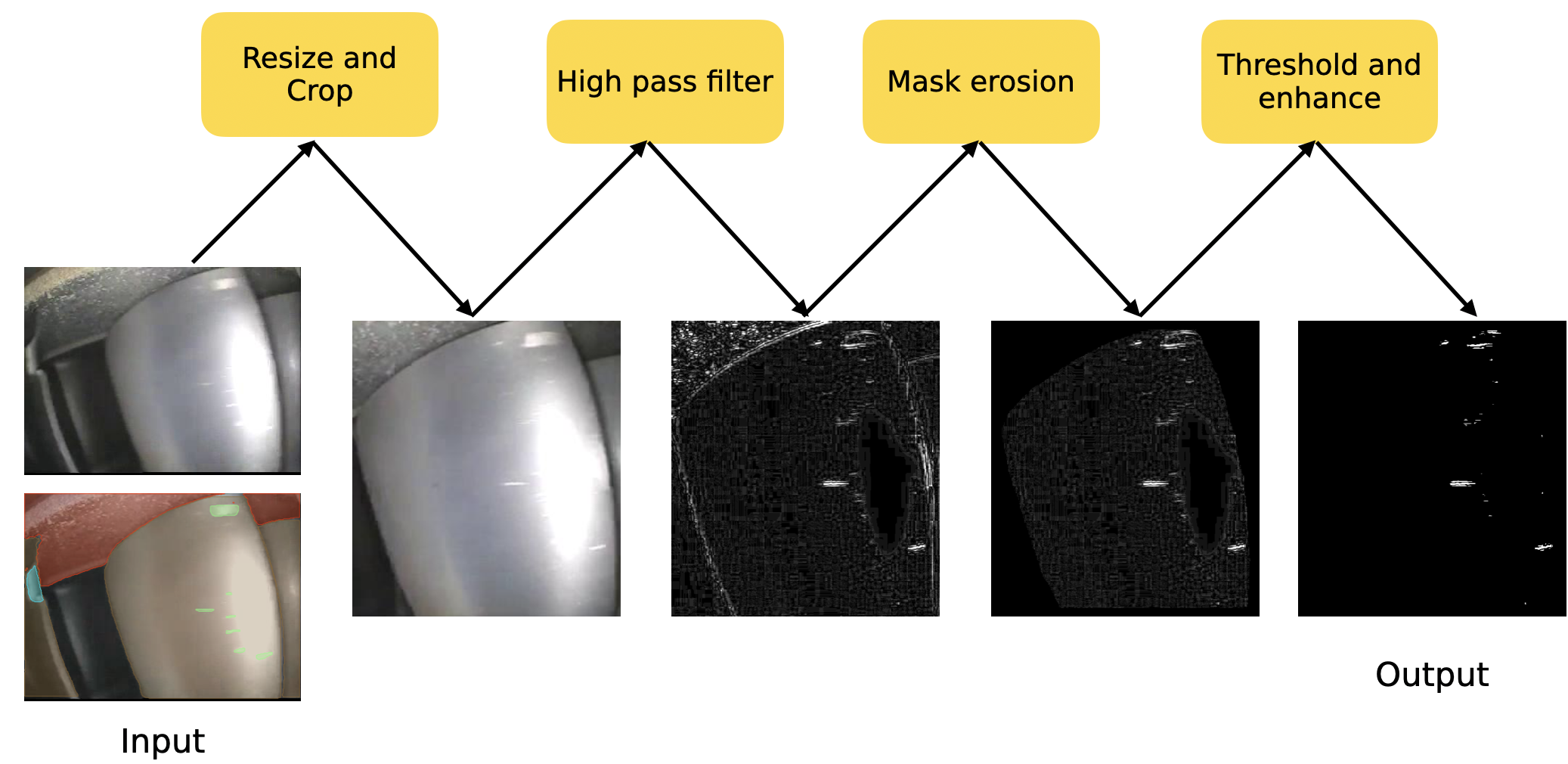}
\caption{Example procedure to generate filtered images that highlight small features on blades leveraging network predictions. Note that enhancing is applied for the final three images for exposition purposes.}
\label{fig:filter_workflow}
\end{figure}

At each frame, the Mask R-CNN needs to detect and mask objects at a wide range of scales. In particular, while larger objects with clear outlines such as blades are more easily detected, smaller patches indicating surface damage on blades are more challenging. In addition, small scratches are difficult to identify through manual inspection of borescope images, which affects the labelling process that generates the training data. Therefore, surface scratches may be missed by the network occasionally. To address this, we suggest the addition of a post-processing step to check the network predictions against filtered images highlighting small scale features. An illustration of this workflow is shown in \Cref{fig:filter_workflow}, and consists of the following steps:
\begin{enumerate}
\item \textbf{Resize and crop} to isolate an individual blade from a frame, using network predictions to find the bounding box of the blade.
\item Apply a \textbf{high-pass filter} to the image within the box. Let $\mathbf{I}$ be the $H\times W$ matrix of grayscale pixel brightness intensities, where $H$ and $W$ are the height and width of the bounding box respectively. The output $\mathbf{I}_h$ is given by
\begin{equation}
\mathbf{I}_h = [\mathbf{I} - \mathbf{I} * \mathbf{G}_\sigma]_+,
\end{equation}
where $\mathbf{G}_\sigma$ is the isotropic Gaussian smoothing kernel with standard deviation $\sigma$ \cite[Ch.~9.12.6]{distante2020image}, $*$ is the (discrete) convolution operator and $[\cdot ]_+$ sets all negative elements of the argument to zero. 
\item \textbf{Mask erosion}: We apply a mask to the output of the high-pass filter which sets the elements outside of the mask to zero. The mask used is a \emph{morphologically eroded} version of the blade mask generated by Mask R-CNN within the bounding box. That is, the expanse of the mask is shrunk at the boundary in the normal direction (see \Cref{fig:erosion}). This operation removes the highlight at the blade edges which is picked up by the high-pass filter.
\item Finally, we apply a \textbf{threshold and enhance} operation to remove pixels with low intensities, and enhance the contrast of the final output.
\end{enumerate}
Using the steps above, small features and discontinuities over the blade are demarcated. 

\begin{figure}
\centering
\begin{subfigure}[b]{.33\textwidth}
\centering
\includegraphics[width=.9\linewidth]{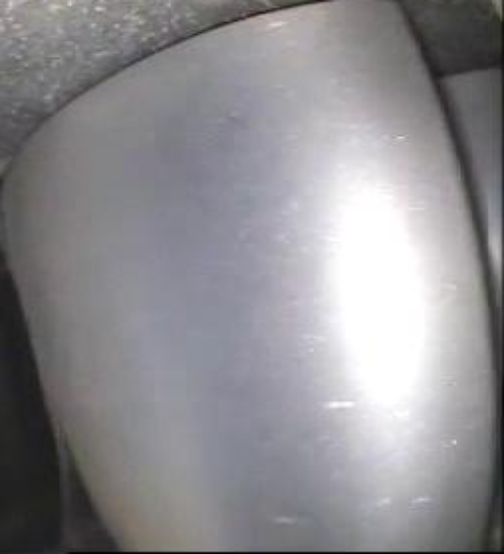}

\vspace{5mm}
\includegraphics[width=.9\linewidth]{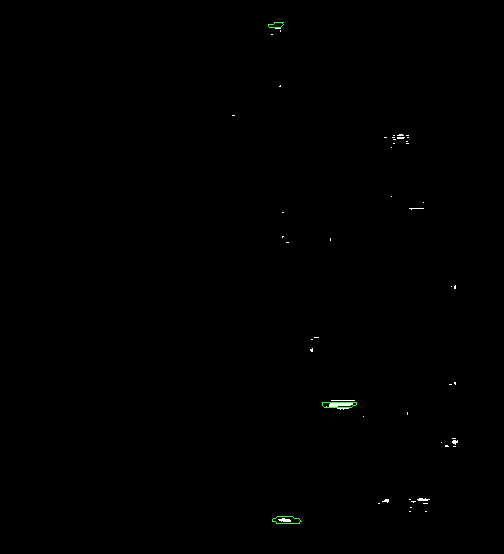}
\end{subfigure}
\begin{subfigure}[b]{.33\textwidth}
\centering
\includegraphics[width=.9\linewidth]{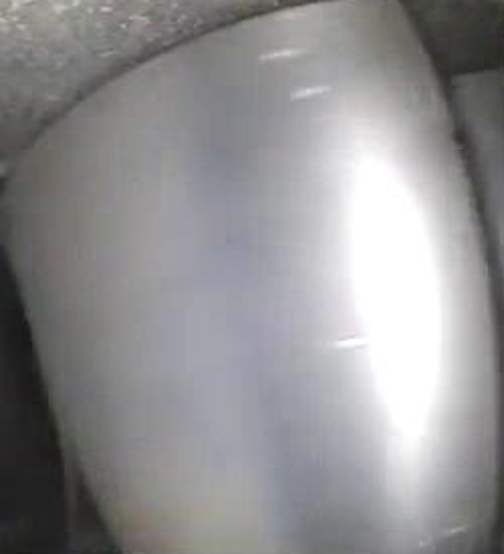}

\vspace{5mm}
\includegraphics[width=.9\linewidth]{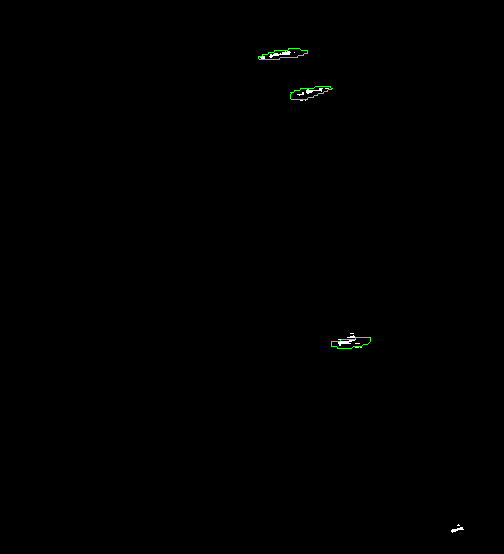}
\end{subfigure}
\begin{subfigure}[b]{.33\textwidth}
\centering
\includegraphics[width=.9\linewidth]{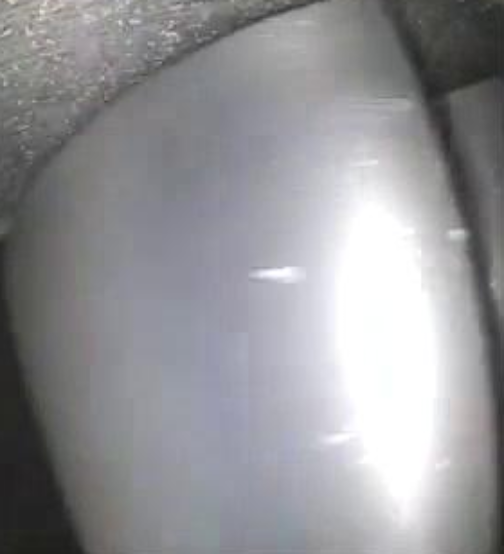}

\vspace{5mm}
\includegraphics[width=.9\linewidth]{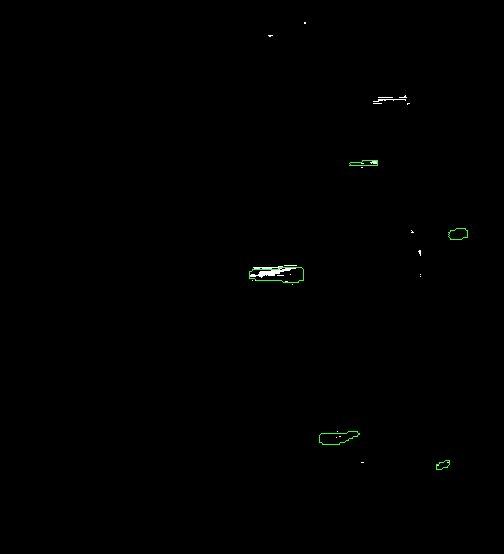}
\end{subfigure}
\caption{Original images (top) and post-processed images (bottom) showing three different blades with surface damage. White pixels indicate the intensity post-filtering, and green contours denote masks from Mask R-CNN. }
\label{fig:hp}
\end{figure}

In \Cref{fig:hp}, these steps are demonstrated on three distinct blades in frames with maximum projected area. It can be seen that the network does not identify all surface scratches, which may even be difficult to identify manually. The high-pass filtering procedure removes the glare of the blade's reflection and shows the presence of small irregularities on the blade surface. It should be noted that this procedure has certain limitations---such as the susceptibility to noise artefacts, the necessity to manually tune the size of the filters, and the inability to distinguish between different types of damage.

%\begin{figure}
%\centering
%\begin{subfigure}[b]{.3\textwidth}
%\begin{minipage}{.5\textwidth}
%\centering
%\includegraphics[width=0.7\linewidth]{figures/orig_24.png}
%\end{minipage}
%\begin{minipage}{.5\textwidth}
%\centering
%\includegraphics[width=0.7\linewidth]{figures/hp_24.png}
%\end{minipage}
%\caption{ }
%\label{subfig:ex0}
%\end{subfigure}
%\begin{subfigure}[b]{.3\textwidth}
%\begin{minipage}{.5\textwidth}
%\includegraphics[width=1\linewidth]{figures/orig_24.png}
%\end{minipage}
%\begin{minipage}{.5\textwidth}
%\includegraphics[width=1\linewidth]{figures/hp_24.png}
%\end{minipage}
%\end{subfigure}
%\caption{HP.}
%\label{fig:hp}
%\end{figure}

\subsection{Analysis of damage around blade row}
\begin{figure}
\centering
\includegraphics[width=0.9\linewidth]{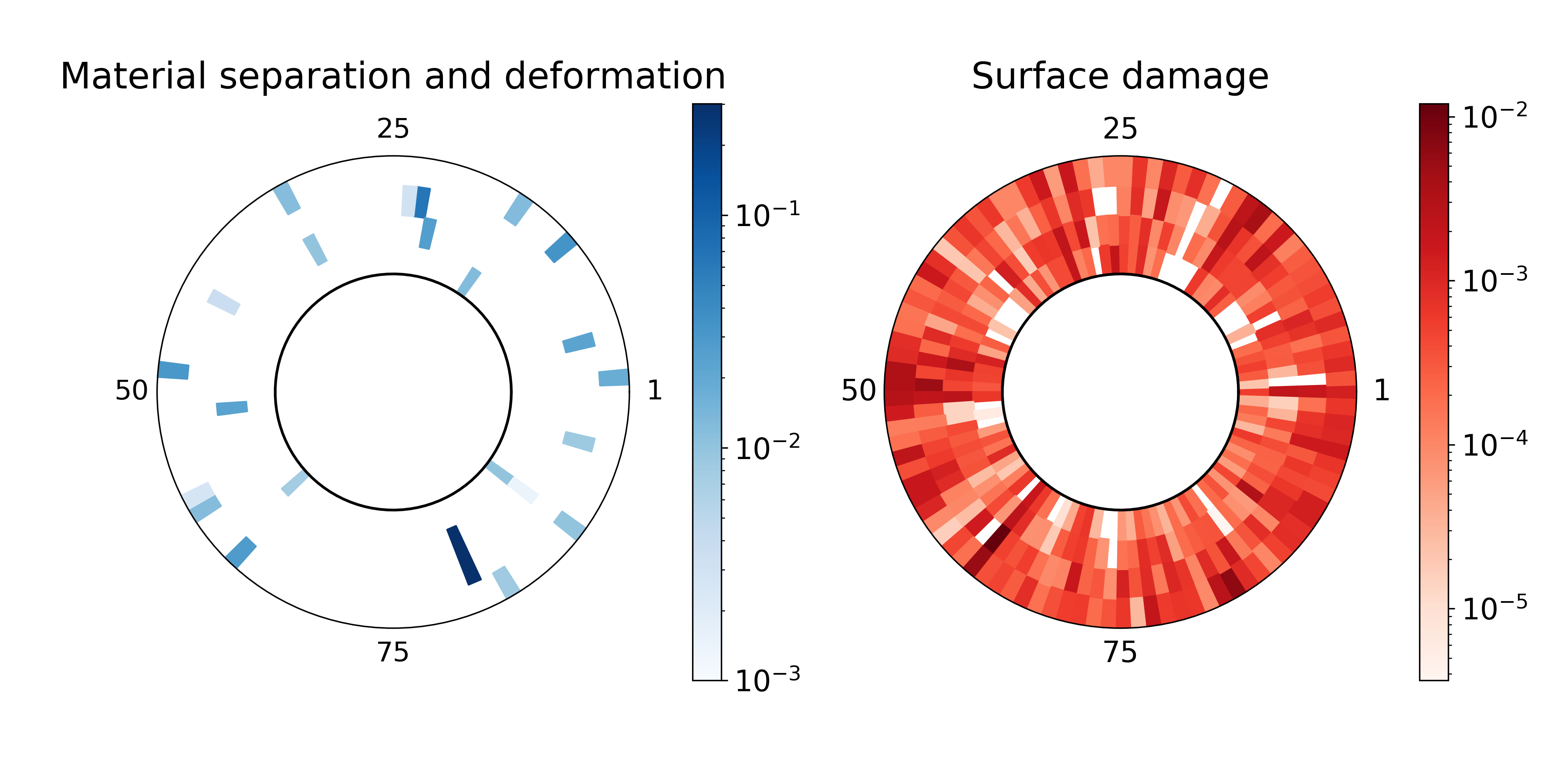}
\caption{Plot of types of damage around a blade row. In both plots, color denotes the extent of damage as a fraction of rotor area, split across four spanwise regions. Spanwise position of the damage is shown with the radial coordinate, and the azimuthal coordinate denotes the rotor number.}
\label{fig:roundplot}
\end{figure}

In the following, we detail the damage found around the rotor row in video 1, where the full blade row is shown. In \Cref{fig:roundplot}, the damage extent on each blade around the rotor disk is summarised and illustrated. For material separation and deformation damage, the extent is calculated from its area in the frame in which it has the maximum projected area; for surface damage, the extent is considered by averaging across all frames, since surface damage is usually made up of a number of small scratches, with only a subset of all surface damage patches on a certain blade visible in a particular frame. In both plots, the amounts of damage are plotted separately for four spanwise regions as a fraction of the entire projected rotor area. These plots reveal important insights about the detected damage. For example, rotor imbalances can be detected when severe separation damage is present in one localised region around the row.

We summarise the output of the entire workflow with an animation accessible at \url{https://youtu.be/nUU-1mpvIpM}. In this video, network prediction labels are visualised along with damage summary plots and a hypothetical performance impact value $\Delta f$, which is a function of the damage types and extent found on each rotor. Provided with hardware enabling fast predictions from the network, this workflow can be deployed in real time during borescope inspections for quick engine sentencing.

%Moreover, one can examine the spatial correlation structure of the detected damages by considering the damage extents on the row as a time series, and find the autocorrelation function of this series. This is defined as
%\begin{equation}
%R_i = \frac{\mathbb{E}_t[X_t X_{t+i}] - \mathbb{E}_t[X_t] \mathbb{E}_t[X_{t+i}] } {\sigma_t[X_t] \sigma_t[X_{t+i}] },
%\end{equation}
%where $i$ is a positive integer denoting the interval between rotors where the correlation is taken across, $X_t$ is the damage extent of the row indexed by $t$, and $\mathbb{E}_t$ and $\sigma_t$ denote the mean and standard deviation across $t$ respectively. The autocorrelation function is evaluated for all three types of damages and plotted in \Cref{fig:autocorr}. Interesting patterns such as the positive correlation of material separation damages between rotors spaced 18 apart from each other and the positive correlations at a number of intervals for material deformations are observed. In general, the autocorrelation function can be used to identify the presence of localised zones of damage.

%\begin{figure}
%\centering
%\includegraphics[width=1\linewidth]{figures/autocorr.png}
%\caption{Autocorrelation function of the damage extent across blades. }
%\label{fig:autocorr}
%\end{figure}

\section{Conclusions and future work}

\begin{figure}
\centering
\includegraphics[width=0.9\linewidth]{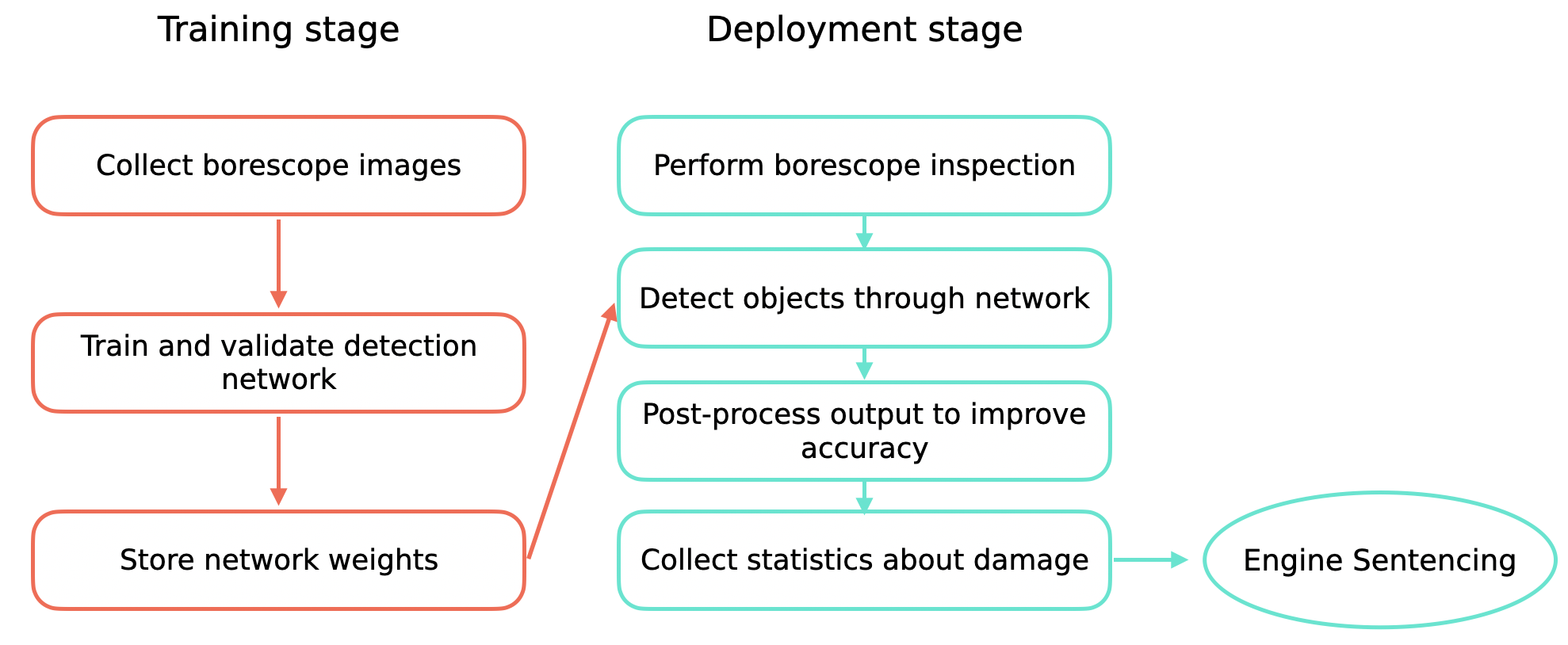}
\caption{Summary of the training and deployment of the automatic borescope inspection framework.}
\label{fig:summary}
\end{figure}

Using methods from computer vision, we show that it is possible to detect damage present on blades via videos collected from borescope inspections. Coupled with algorithms for blade tracking, this enables an automatic workflow that collects statistics about the types and extent of damage found on rotor blades. To summarise, the steps for training and deploying the automatic borescope inspection framework in practice are outlined in \Cref{fig:summary}.

To take this work further, we need to convert the collected damage statistics over several frames to a full assessment of the damaged geometry of a blade. This involves the mapping of detected damage in 2D to a 3D model of a blade in CAD representation. The 3D blade models serve as digital twins that are amenable to aerodynamic analysis to assess the implications of the damage. For example, in \cite{wong2021blade,wong2020blade}, the authors propose a method to measure a metric distance of a geometry from a collection of blades that are known to adhere to performance and safety standards---called a blade envelope. Integrating the present framework with the workflow in \cite{wong2021blade,wong2020blade}, we envision that our system can serve as a component in a real-time assessment and recommendation system to aid inspectors in engine sentencing and engineers in designing robust bladed components.

\section*{Acknowledgements}
CYW acknowledges financial support from the Cambridge Trust, Jesus College, Cambridge, and the Data-Centric Engineering programme of The Alan Turing Institute. PS acknowledges the financial support of Rolls-Royce plc, and the Strategic Priorities Fund---delivered by UK Research and Innovation, with this award managed by EPSRC (EP/T001569/1). This research was supported in part through computational resources provided by The Alan Turing Institute and with the help of a generous gift from Microsoft Corporation.

\bibliographystyle{plain}
\bibliography{aiaa}

\begin{thebibliography}{10}

\bibitem{adamczuk2013early}
Rafael~R. Adamczuk and Joerg~R. Seume.
\newblock Early {Assessment} of {Defects} and {Damage} in {Jet} {Engines}.
\newblock {\em Procedia CIRP}, 11:328--333, January 2013.

\bibitem{aust2019taxonomy}
Jonas Aust and Dirk Pons.
\newblock Taxonomy of {Gas} {Turbine} {Blade} {Defects}.
\newblock {\em Aerospace}, 6(5):58, May 2019.

\bibitem{bian2016multiscale}
Xiao Bian, Ser~Nam Lim, and Ning Zhou.
\newblock Multiscale fully convolutional network with application to industrial
  inspection.
\newblock In {\em 2016 {IEEE} {Winter} {Conference} on {Applications} of
  {Computer} {Vision} ({WACV})}, pages 1--8, Lake Placid, NY, USA, March 2016.
  IEEE.

\bibitem{carter2005common}
Tim~J Carter.
\newblock Common failures in gas turbine blades.
\newblock {\em Engineering Failure Analysis}, 12(2):237--247, April 2005.

\bibitem{cordts2016cityscapes}
Marius Cordts, Mohamed Omran, Sebastian Ramos, Timo Rehfeld, Markus Enzweiler,
  Rodrigo Benenson, Uwe Franke, Stefan Roth, and Bernt Schiele.
\newblock The {Cityscapes} {Dataset} for {Semantic} {Urban} {Scene}
  {Understanding}.
\newblock In {\em 2016 {IEEE} {Conference} on {Computer} {Vision} and {Pattern}
  {Recognition} ({CVPR})}, pages 3213--3223, Las Vegas, NV, USA, June 2016.
  IEEE.

\bibitem{distante2020image}
Arcangelo Distante and Cosimo Distante.
\newblock Image {Enhancement} {Techniques}.
\newblock In Arcangelo Distante and Cosimo Distante, editors, {\em Handbook of
  {Image} {Processing} and {Computer} {Vision}: {Volume} 1: {From} {Energy} to
  {Image}}, pages 387--484. Springer International Publishing, Cham, 2020.

\bibitem{drury2001human}
Colin~G Drury and Jean Watson.
\newblock Human {Factors} {Good} {Practices} {In} {Borescope} {Inspection}.
\newblock Technical report, Federal Aviation Administration, May 2001.

\bibitem{federal_aviation_administration2013aviation}
{Federal Aviation Administration}.
\newblock {\em Aviation {Maintenance} {Technician} {Handbook} - {General}
  ({FAA} {Handbooks})}.
\newblock Aviation Supplies and Academics, Inc., March 2013.

\bibitem{girshick2015fast}
Ross Girshick.
\newblock Fast {R}-{CNN}.
\newblock In {\em Proceedings of the 2015 {IEEE} {International} {Conference}
  on {Computer} {Vision} ({ICCV})}, {ICCV} '15, pages 1440--1448, USA, December
  2015. IEEE Computer Society.

\bibitem{girshick2014rich}
Ross Girshick, Jeff Donahue, Trevor Darrell, and Jitendra Malik.
\newblock Rich {Feature} {Hierarchies} for {Accurate} {Object} {Detection} and
  {Semantic} {Segmentation}.
\newblock In {\em 2014 {IEEE} {Conference} on {Computer} {Vision} and {Pattern}
  {Recognition}}, pages 580--587, Columbus, OH, USA, June 2014. IEEE.

\bibitem{hanschke2017effect}
Benjamin Hanschke, Thomas Klauke, and Arnold Kühhorn.
\newblock {\em The {Effect} of {Foreign} {Object} {Damage} on {Compressor}
  {Blade} {High} {Cycle} {Fatigue} {Strength}}, volume Volume 7A: Structures
  and Dynamics of {\em Turbo {Expo}: {Power} for {Land}, {Sea}, and {Air}}.
\newblock 2017.
\newblock \_eprint:
  https://asmedigitalcollection.asme.org/GT/proceedings-pdf/GT2017/50923/V07AT31A005/2434362/v07at31a005-gt2017-63599.pdf.

\bibitem{he2017mask}
K.~He, G.~Gkioxari, P.~Dollár, and R.~Girshick.
\newblock Mask {R}-{CNN}.
\newblock In {\em 2017 {IEEE} {International} {Conference} on {Computer}
  {Vision} ({ICCV})}, pages 2980--2988, October 2017.
\newblock ISSN: 2380-7504.

\bibitem{khani2012towards}
Nqobile Khani, Clara Segovia, Rukshan Navaratne, Vishal Sethi, Riti Singh, and
  Pericles Pilidis.
\newblock Towards {Development} of a {Diagnostic} and {Prognostic} {Tool} for
  {Civil} {Aero}-{Engine} {Component} {Degradation}.
\newblock In {\em {GTINDIA2012}}, pages 803--814, ASME 2012 Gas Turbine India
  Conference, December 2012.

\bibitem{kirillov2020pointrend}
Alexander Kirillov, Yuxin Wu, Kaiming He, and Ross Girshick.
\newblock {PointRend}: {Image} {Segmentation} {As} {Rendering}.
\newblock In {\em 2020 {IEEE}/{CVF} {Conference} on {Computer} {Vision} and
  {Pattern} {Recognition} ({CVPR})}, pages 9796--9805, Seattle, WA, USA, June
  2020. IEEE.

\bibitem{krizhevsky2012imagenet}
Alex Krizhevsky, Ilya Sutskever, and Geoffrey~E. Hinton.
\newblock {ImageNet} {Classification} with {Deep} {Convolutional} {Neural}
  {Networks}.
\newblock {\em Advances in Neural Information Processing Systems},
  25:1097--1105, 2012.

\bibitem{lecun1998gradient-based}
Y.~Lecun, L.~Bottou, Y.~Bengio, and P.~Haffner.
\newblock Gradient-based learning applied to document recognition.
\newblock {\em Proceedings of the IEEE}, 86(11):2278--2324, November 1998.

\bibitem{li2017fully}
Y.~Li, H.~Qi, J.~Dai, X.~Ji, and Y.~Wei.
\newblock Fully {Convolutional} {Instance}-{Aware} {Semantic} {Segmentation}.
\newblock In {\em 2017 {IEEE} {Conference} on {Computer} {Vision} and {Pattern}
  {Recognition} ({CVPR})}, pages 4438--4446, July 2017.
\newblock ISSN: 1063-6919.

\bibitem{lin2017feature}
Tsung-Yi Lin, Piotr Dollar, Ross Girshick, Kaiming He, Bharath Hariharan, and
  Serge Belongie.
\newblock Feature {Pyramid} {Networks} for {Object} {Detection}.
\newblock In {\em 2017 {IEEE} {Conference} on {Computer} {Vision} and {Pattern}
  {Recognition} ({CVPR})}, pages 936--944, Honolulu, HI, July 2017. IEEE.

\bibitem{lin2014microsoft}
Tsung-Yi Lin, Michael Maire, Serge Belongie, James Hays, Pietro Perona, Deva
  Ramanan, Piotr Dollár, and C.~Lawrence Zitnick.
\newblock Microsoft {COCO}: {Common} {Objects} in {Context}.
\newblock In David Fleet, Tomas Pajdla, Bernt Schiele, and Tinne Tuytelaars,
  editors, {\em Computer {Vision} – {ECCV} 2014}, Lecture {Notes} in
  {Computer} {Science}, pages 740--755, Cham, 2014. Springer International
  Publishing.

\bibitem{long2015fully}
J.~Long, E.~Shelhamer, and T.~Darrell.
\newblock Fully convolutional networks for semantic segmentation.
\newblock In {\em 2015 {IEEE} {Conference} on {Computer} {Vision} and {Pattern}
  {Recognition} ({CVPR})}, pages 3431--3440, June 2015.
\newblock ISSN: 1063-6919.

\bibitem{meher-homji1998gas}
Cyrus~B. Meher-Homji and George Gabriles.
\newblock Gas {Turbine} {Blade} {Failures} - {Causes}, {Avoidance}, {And}
  {Troubleshooting}.
\newblock Technical report, Texas A\&M University. Turbomachinery Laboratories,
  1998.

\bibitem{padilla2020survey}
R.~Padilla, S.~L. Netto, and E.~A. B.~da Silva.
\newblock A {Survey} on {Performance} {Metrics} for {Object}-{Detection}
  {Algorithms}.
\newblock In {\em 2020 {International} {Conference} on {Systems}, {Signals} and
  {Image} {Processing} ({IWSSIP})}, pages 237--242, July 2020.
\newblock ISSN: 2157-8702.

\bibitem{paszke2019pytorch}
Adam Paszke, Sam Gross, Francisco Massa, Adam Lerer, James Bradbury, Gregory
  Chanan, Trevor Killeen, Zeming Lin, Natalia Gimelshein, Luca Antiga, Alban
  Desmaison, Andreas Kopf, Edward Yang, Zachary DeVito, Martin Raison, Alykhan
  Tejani, Sasank Chilamkurthy, Benoit Steiner, Lu~Fang, Junjie Bai, and Soumith
  Chintala.
\newblock {PyTorch}: {An} {Imperative} {Style}, {High}-{Performance} {Deep}
  {Learning} {Library}.
\newblock In H.~Wallach, H.~Larochelle, A.~Beygelzimer,
  F.~d{\textbackslash}textquotesingle Alché-Buc, E.~Fox, and R.~Garnett,
  editors, {\em Advances in {Neural} {Information} {Processing} {Systems} 32},
  pages 8024--8035. Curran Associates, Inc., 2019.

\bibitem{pinheiro2015learning}
Pedro~O. Pinheiro, Ronan Collobert, and Piotr Dollár.
\newblock Learning to segment object candidates.
\newblock In {\em Proceedings of the 28th {International} {Conference} on
  {Neural} {Information} {Processing} {Systems} - {Volume} 2}, {NIPS}'15, pages
  1990--1998, Montreal, Canada, December 2015. MIT Press.

\bibitem{redmon2016you}
J.~Redmon, S.~Divvala, R.~Girshick, and A.~Farhadi.
\newblock You {Only} {Look} {Once}: {Unified}, {Real}-{Time} {Object}
  {Detection}.
\newblock In {\em 2016 {IEEE} {Conference} on {Computer} {Vision} and {Pattern}
  {Recognition} ({CVPR})}, pages 779--788, June 2016.
\newblock ISSN: 1063-6919.

\bibitem{ren2017faster}
Shaoqing Ren, Kaiming He, Ross Girshick, and Jian Sun.
\newblock Faster {R}-{CNN}: {Towards} {Real}-{Time} {Object} {Detection} with
  {Region} {Proposal} {Networks}.
\newblock {\em IEEE Transactions on Pattern Analysis and Machine Intelligence},
  39(6):1137--1149, June 2017.

\bibitem{shen2019deep}
Zejiang Shen, Xili Wan, Feng Ye, Xinjie Guan, and Shuwen Liu.
\newblock Deep {Learning} based {Framework} for {Automatic} {Damage}
  {Detection} in {Aircraft} {Engine} {Borescope} {Inspection}.
\newblock In {\em 2019 {International} {Conference} on {Computing},
  {Networking} and {Communications} ({ICNC})}, pages 1005--1010, Honolulu, HI,
  USA, February 2019. IEEE.

\bibitem{tanner2005method}
Joseph Tanner and Mark Bailey.
\newblock Method and system for automated repair design of damaged blades of a
  compressor or turbine.
\newblock U.S. Patent US20050033555A1, February 2005.

\bibitem{taylor2020predicting}
J.~V. Taylor, B.~Conduit, A.~Dickens, C.~Hall, M.~Hillel, and R.~J. Miller.
\newblock Predicting the {Operability} of {Damaged} {Compressors} {Using}
  {Machine} {Learning}.
\newblock {\em Journal of Turbomachinery}, 142(051010), April 2020.

\bibitem{wong2021blade}
Chun~Yui Wong, Pranay Seshadri, Ashley Scillitoe, Andrew~B. Duncan, and
  Geoffrey Parks.
\newblock Blade {Envelopes} {Part} {I}: {Concept} and {Methodology}.
\newblock {\em arXiv:2011.11636 [cs, stat]}, January 2021.
\newblock arXiv: 2011.11636.

\bibitem{wong2020blade}
Chun~Yui Wong, Pranay Seshadri, Ashley Scillitoe, Bryn~Noel Ubald, Andrew~B.
  Duncan, and Geoffrey Parks.
\newblock Blade {Envelopes} {Part} {II}: {Multiple} {Objectives} and {Inverse}
  {Design}.
\newblock {\em arXiv:2012.15579 [cs]}, December 2020.
\newblock arXiv: 2012.15579.

\bibitem{wu2019detectron2}
Yuxin Wu, Alexander Kirillov, Francisco Massa, Wan-Yen Lo, and Ross Girshick.
\newblock {\em Detectron2}.
\newblock 2019.

\end{thebibliography}

\end{document}